\documentclass[10pt,twocolumn,letterpaper]{article}

\usepackage[pagenumbers]{wacv} 

\usepackage[skip=3pt]{caption}

\usepackage{tikz,tikzscale}
\usepackage{pgfplots}
\pgfplotsset{compat=1.18}

\usepackage{listings}
\usepackage[most]{tcolorbox}  
\newtcolorbox[auto counter, number within=section]{instructionbox}[2][]{%
  colback=gray!5!white,
  colframe=gray!50!black,
  fonttitle=\small,
  fontupper=\small,
  title=Box~\thetcbcounter: #2,
  #1
}

\usepackage[ruled,vlined]{algorithm2e}
\def\para#1{\medskip\noindent\textbf{#1}}

\definecolor{wacvblue}{rgb}{0.21,0.49,0.74}
\usepackage[pagebackref,breaklinks,colorlinks,allcolors=wacvblue]{hyperref}

\def\wacvPaperID{***} 
\def\confName{WACV}
\def\confYear{2026}

\title{LAMS-Edit: \underline{L}atent and \underline{A}ttention \underline{M}ixing with \underline{S}chedulers for Improved \\ Content Preservation in Diffusion-Based Image and Style \underline{Edit}ing}

\author{Wingwa Fu \quad Takayuki Okatani\\
Graduate School of Information Sciences, Tohoku University\\
{\tt\small fu.wingwa.r8@dc.tohoku.ac.jp, okatani@vision.is.tohoku.ac.jp}
}

\begin{document}

\twocolumn[{
\renewcommand\twocolumn[1][]{#1}
\maketitle
\begin{center}
    \centering
    \captionsetup{type=figure}
    \includegraphics[width=\textwidth]{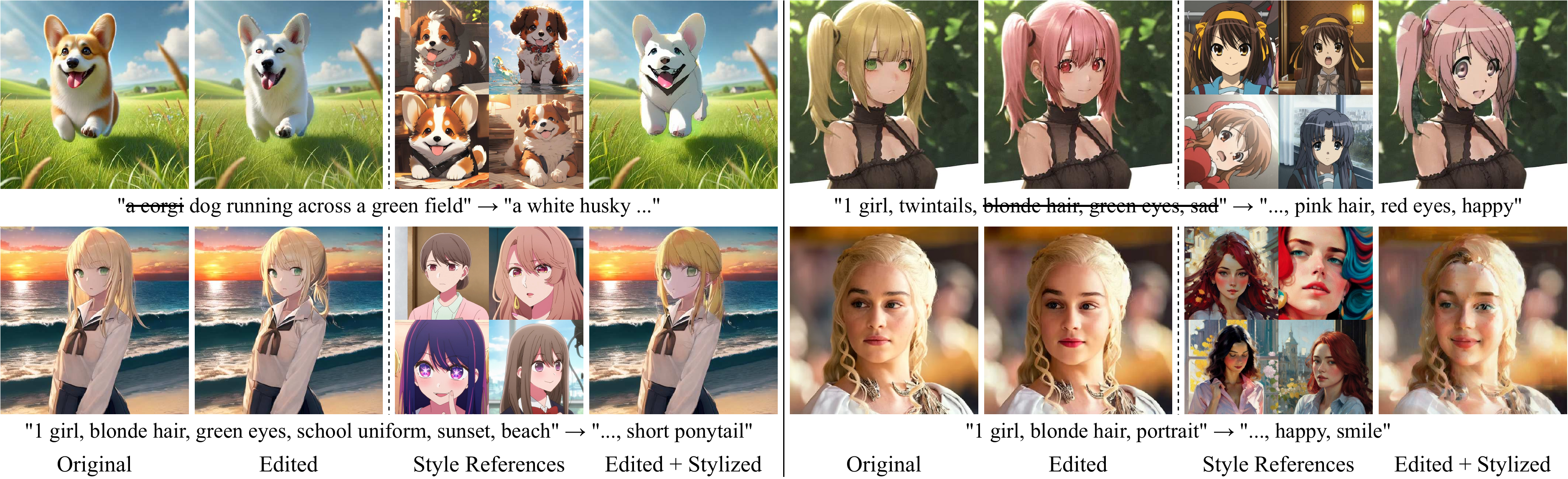}
    \captionof{figure}{Overview of LAMS-Edit's capabilities. LAMS-Edit enhances structure and content preservation in T2I editing and style transfer.}
    \label{fig:teaser}
\end{center}
}]

\begin{abstract}
Text-to-Image editing using diffusion models faces challenges in balancing content preservation with edit application and handling real-image editing. To address these, we propose LAMS-Edit, leveraging intermediate states from the inversion process—an essential step in real-image editing—during edited image generation. Specifically, latent representations and attention maps from both processes are combined at each step using weighted interpolation, controlled by a scheduler. This technique, Latent and Attention Mixing with Schedulers (LAMS), integrates with Prompt-to-Prompt (P2P) to form LAMS-Edit—an extensible framework that supports precise editing with region masks and enables style transfer via LoRA. Extensive experiments demonstrate that LAMS-Edit effectively balances content preservation and edit application.
\end{abstract}    
\section{Introduction}
\label{sec:intro}

Image editing using diffusion models has gained increasing attention due to its potential in professional workflows, such as digital art, content creation, and advertising. While text-to-image (T2I) generation \cite{ho2020ddpm, song2020ddim, rombach2021stablediffusion} enables users to create images from natural language descriptions, real-world applications often require modifying existing images rather than generating new ones from scratch. This need has driven research into both content and style editing.

Various approaches have been explored to enable content editing, which involves adding or removing specific objects, modifying shapes, or making local adjustments within an image. In this process, users provide instructions through text prompts, mask images, and other input methods. The goal is to apply edits as intended while preserving the image's structure and semantic content. Some methods utilize mask-based inpainting for localized edits \cite{Avrahami_2022_CVPR,avrahami2023blendedlatent,couairon2023diffedit,simsar2023lime,wang2023instructedit,park2024shape,cao_2023_masactrl}, while others leverage internal representations such as latent features and attention maps \cite{hertz2022prompt,epstein2023diffusion,parmar2023pix2pixzero,Tumanyan_2023_CVPR} or external resources like reference images \cite{yang2022paint,zhang2023adding} to guide modifications. Additionally, some methods rely on fine-tuning \cite{kawar2023imagic,brooks2022instructpix2pix,yang2022paint,zhang2023adding} or parameter optimization of diffusion models \cite{mokady2022null,Wu2022UncoveringTD,chen2024tino,dong2023prompt}. However, approaches that do not require fine-tuning or optimization \cite{hertz2022prompt,epstein2023diffusion,parmar2023pix2pixzero,Tumanyan_2023_CVPR,meng2022sdedit,mao2023Guided,couairon2023diffedit,simsar2023lime,wang2023instructedit,park2024shape,cao_2023_masactrl,brack2023leditspp,ju2023pnpinv} have recently gained greater attention due to their computational efficiency.

Various methods have also been explored for style transfer, with significant advancements in fine-tuning techniques \cite{ruiz2022dreambooth,gal2022textual,hu2022lora}. Users specify the desired transformation by providing reference images or text prompts. As with content editing, maintaining the structure and semantic content of an image is a fundamental requirement for style transfer \cite{huang2024diffstyler,li2024diffstyler,Zhang_2023_inst,cheng2023general}. 

Despite extensive research in these areas, existing methods for content editing and style transfer still often struggle to achieve satisfactory performance. Achieving precise edits as intended while preserving the original image’s structure and semantic content remains a challenge, and this difficulty is particularly pronounced in real-image editing.

In this study, we aim to address this challenge by leveraging the intermediate steps of the inversion process, which is commonly used for real-image editing, and utilizing this information to extend P2P (Prompt-to-Prompt) \cite{hertz2022prompt}. 
Inverting the image generation process of a diffusion model—most notably through DDIM inversion \cite{song2020ddim}—allows us to obtain an initial latent variable that serves as the starting point for reconstructing the image using the diffusion model. The latent obtained through inversion is considered ``correct'' in the sense that the reconstructed image is nearly identical to the original\footnote{Here, we assume both inversion and generation are performed under conditional generation, where text prompts are provided as conditioning}. However, this latent differs from the standard initial latent used for generation from scratch, which consists of pure noise \cite{mokady2022null,huang2024dualscheduleinv,samuel2025fixedpointinv}. As a result, applying P2P directly to an inversion-derived latent fails to produce satisfactory results \cite{wallace2022edict}.

We propose a novel approach that utilizes not only the final inversion result—the initial latent—but also its trajectory, i.e., the intermediate steps of the inversion process, to generate edited images. This contrasts with conventional methods that rely solely on the inversion-derived initial latent. We hypothesize that the initial latent alone does not fully capture essential information from the original image; instead, critical structural and fine-grained details are embedded in the intermediate steps of the inversion process.

We demonstrate that a simple linear combination of inversion-derived latents and attention maps with their counterparts during the generation process yields strong results. This effect is further enhanced when combined with a scheduling strategy that assigns higher weights to inversion-derived latents and attention maps in the early stages of generation (i.e., denoising), gradually reducing their influence in later stages. We name this approach as LAMS (Latent and Attention Mixing with Schedulers) and propose LAMS-Edit, a framework that integrates LAMS with P2P. 

LAMS-Edit is an image editing method that does not require fine-tuning or optimization. It allows for enhanced spatial precision by optionally specifying an editing region mask. Furthermore, it seamlessly integrates style transfer using LoRA \cite{hu2022lora}, enabling the simultaneous application of both content editing and style transfer.

\section{Related Work}
\label{sec:related_work}

\subsection{Text-to-Image Editing with Diffusion Models}

\para{Fine-Tuning-Based Approaches.} Some approaches adapt pre-trained diffusion models for text-guided image editing. Imagic \cite{kawar2023imagic} fine-tunes the model while optimizing text embeddings to align input images with target descriptions. InstructPix2Pix \cite{brooks2022instructpix2pix} trains Stable Diffusion (SD) on image-instruction pairs to enable text-driven modifications. Paint by Example \cite{yang2022paint} facilitates exemplar-based editing using a CLIP-based classifier. ControlNet \cite{zhang2023adding} trains an auxiliary network to process visual guidance, such as edges and depth maps, for more controlled editing. SINE \cite{zhang2022sine} employs patch-based fine-tuning and extends classifier-free guidance with model-based guidance for image editing. Text2LIVE \cite{bar2022text2live} trains a generator to produce an RGBA edit layer for localized text-driven edits in images and videos. While these methods enable effective edits, they require substantial computational resources.

\para{Optimization-Based Approaches.} Instead of fine-tuning a base model, other approaches refine inputs at inference time, eliminating the need for retraining. Null-Text Inversion \cite{mokady2022null} optimizes unconditional embeddings to improve reconstruction and enable further text-guided modifications. DiffusionDisentanglement \cite{Wu2022UncoveringTD} separates text embeddings into neutral and styled components, allowing controlled attribute adjustments. TiNO-Edit \cite{chen2024tino} refines noise patterns and diffusion steps to maintain image consistency during edits. Prompt Tuning (PT) \cite{dong2023prompt} refines the embedding of the original image prompt to cope with the inaccuracy in DDIM inversion. Specifically, it optimizes the embedding at each timestep to ensure that, when reconstructing the original image from the inverted initial latent, the latent remains close to the inversion-derived latent; the optimized embedding is then interpolated with that of the target prompt during the generation process. Although PT shares similarities with our method in that it leverages the latent representations from the inversion process, it interpolates prompt embeddings, making its technical approach and objective distinct.

\para{Tuning-Free Approaches.} Recent methods enable efficient editing by manipulating internal representations without fine-tuning or optimization. Prompt-to-Prompt \cite{hertz2022prompt} and Diffusion Self-Guidance \cite{epstein2023diffusion} modify attention maps for localized control and attribute adjustments. Pix2Pix-Zero \cite{parmar2023pix2pixzero} and Plug-and-Play \cite{Tumanyan_2023_CVPR} leverage cross-attention and deep features for content preservation, while SDEdit \cite{meng2022sdedit} and Guided Image Synthesis \cite{mao2023Guided} refine edits through noise injection and latent manipulation, respectively. 
PnPInversion \cite{ju2023pnpinv} separates the editing into source and target branches and guides the process by adding and subtracting latent variables between them. EDICT \cite{wallace2022edict} reformulates DDIM to improve inversion, while LEDITS++ \cite{brack2023leditspp} utilizes a higher-order differential equation solver to achieve more accurate inversion and combines this with text-driven editing. On the other hand, 
GLIGEN \cite{li2023gligen} integrates grounding inputs for spatial control. Mask-based methods, such as Blended Diffusion \cite{Avrahami_2022_CVPR}, Blended Latent Diffusion \cite{avrahami2023blendedlatent}, and Shape-Guided Diffusion \cite{park2024shape}, rely on manual masks, while DiffEdit \cite{couairon2023diffedit}, LIME \cite{simsar2023lime}, MasaCtrl \cite{cao_2023_masactrl}, and LEDITS++ \cite{brack2023leditspp} generate masks from internal representations. InstructEdit \cite{wang2023instructedit} uses ChatGPT and SAM for automated mask generation. While efficient and effective, these methods often struggle to balance content preservation with intended modifications.

\subsection{Style Transfer with Diffusion Models}

\para{Personalization Techniques.} To adapt the model to a specific style domain, some methods fine-tune it to learn particular styles from limited data. DreamBooth \cite{ruiz2022dreambooth} adapts diffusion models to a given style using a few reference images. Textual Inversion \cite{gal2022textual} embeds novel concepts into the text space, enabling style- or object-specific generation via learned tokens. LoRA (Low-Rank Adaptation) \cite{hu2022lora} provides a more efficient alternative by fine-tuning only a subset of model weights for style adaptation. While these methods allow diffusion models to generate images in a learned style, they are not inherently designed for style transfer. However, they serve as the foundation for subsequent style editing research.

\para{Style Editing Methods.} Building on these techniques, later approaches enable style transfer while preserving the content. DiffStyler \cite{huang2024diffstyler,li2024diffstyler} employs dual-diffusion architectures to maintain structural integrity. InST \cite{Zhang_2023_inst} uses an attention-based textual inversion approach to extract and transfer high-level artistic attributes. Similarly, VCT \cite{cheng2023general} enables image-to-image translation by preserving content while incorporating style from a reference image through dual-stream denoising.

\section{Preliminaries}
\label{sec:preliminaries}

\subsection{Stable Diffusion}
Our research builds upon Stable Diffusion (SD) \cite{rombach2021stablediffusion}, a Diffusion Model (DM) that operates in a lower-dimensional latent space rather than pixel space. Given an input image $\mathbf{x}_0$, the encoder $\mathcal{E}$ maps it to the latent space as $\mathbf{z}_0 = \mathcal{E}(\mathbf{x}_0)$. Diffusion processes are performed in the latent space, generating a latent variable $\mathbf{\hat{z}}_0$, which is then passed to the decoder $\mathcal{D}$ to generate the image as $\mathbf{\hat{x}}_0 = \mathcal{D}(\mathbf{\hat{z}}_0)$.

In the generation (i.e., denoising) process, a U-Net architecture is used to predict the noise $\boldsymbol{\epsilon}_\theta$ at each step, where self-attention and cross-attention mechanisms play a crucial role. The attention maps are computed as:
\begin{equation}
\mathbf{A} = \text{softmax}\left(\frac{\mathbf{q} \mathbf{k}^\top}{\sqrt{d_k}}\right),
\label{eq:attention}
\end{equation}
where queries ($\mathbf{q}$) and keys ($\mathbf{k}$) are defined as:
\begin{equation}
\mathbf{q} = \mathbf{W}_q \cdot \mathbf{z}_t, \quad \mathbf{k} = 
\begin{cases} 
\mathbf{W}_k \cdot \mathbf{z}_t & \text{(self-attention)} \\
\mathbf{W}_k \cdot \tau & \text{(cross-attention)} 
\end{cases},
\label{eq:attention_qk}
\end{equation}
where $\mathbf{W}_q$ and $\mathbf{W}_k$ are learned projection matrices, and $\tau$ represents the embedding of an input textual prompt used to guide the image generation. Manipulating the attention maps allows for the control of content generation \cite{hertz2022prompt, liu2024freepromptedit}.

\subsection{Prompt-to-Prompt}
Prompt-to-Prompt (P2P) \cite{hertz2022prompt} leverages the attention mechanisms in diffusion models to enable T2I editing by modifying the text prompt. It refines images by adjusting cross-attention maps corresponding to the modified textual prompt. By replacing or adjusting attention activations, it selectively modifies only the image regions associated with the edited tokens while preserving the overall structure. Since P2P is designed for editing generated images, an inversion technique is required to enable real image editing.

\subsection{DDIM Inversion}

DDIM inversion \cite{song2020ddim} is the most widely used method for inverting the denoising process in diffusion models, particularly for real image editing. Since precisely inverting the denoising process is challenging, DDIM inversion introduces an approximation to simplify the computation. While the reconstruction of the original image from the obtained initial latent is generally effective, this approximation causes the reconstruction process to deviate from genuine image generation, which starts from pure noise \cite{mokady2022null,huang2024dualscheduleinv,samuel2025fixedpointinv}. This discrepancy may be a key factor that makes real image editing more challenging. The proposed method aims to address this issue, as explained below.

\section{Method}
\label{sec:method}

\begin{figure}[t]
    \centering
    \includegraphics[width=\linewidth]{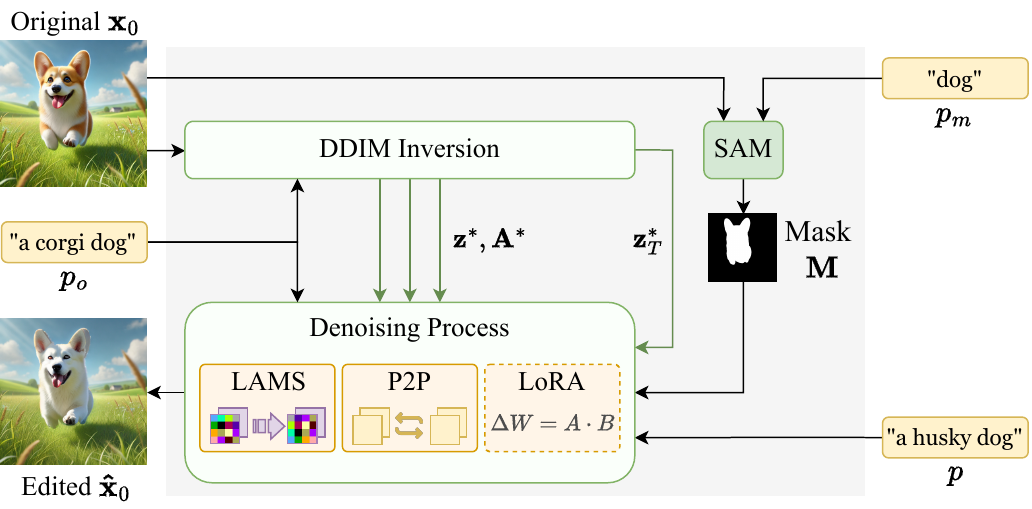}
    \captionof{figure}{
    Overview of LAMS-Edit. Given an input image $\mathbf{x}_0$, DDIM inversion computes the initial latent representation $\mathbf{z}_T^*$, along with intermediate latent representations $\mathbf{z}^*$ and attention maps $\mathbf{A}^*$. These are then utilized by LAMS in the generation process, which is integrated with P2P to produce the edited image $\mathbf{\hat{x}}_0$. Optionally, a mask $\mathbf{M}$ generated by SAM \cite{kirillov2023sam} can be applied to improve the spatial precision of edits. Additionally, LoRA-based style transfer can be applied simultaneously.}
    \label{fig:lamsedit_overview_real_img}
\end{figure}

This section introduces LAMS-Edit, a tuning-free and optimization-free framework for T2I editing, which extends Prompt-to-Prompt (P2P) \cite{hertz2022prompt}; see Fig.~\ref{fig:lamsedit_overview_real_img}. The core component is Latent and Attention Mixing with Schedulers (LAMS) (Sec.~\ref{sec:lam} and \ref{sec:lams_schedulers}). Optionally, a mask can be specified to enhance the spatial accuracy of edits (Sec.~\ref{sec:lamsedit_sam_masking}), and LoRA-based style transfer can also be incorporated (Sec.~\ref{sec:lamsedit_style_transfer}).

\subsection{Overview}

As shown in Fig.~\ref{fig:lamsedit_overview_real_img}, LAMS-Edit takes as input a single image $\mathbf{x}_0$, an original prompt $p_o$\footnote{This can be either provided alongside $\mathbf{x}_0$, specified by the user, or generated from $\mathbf{x}_0$ through image captioning.}, and a target prompt $p$. Optionally, a mask $\mathbf{M}$ can be used, in which case the user specifies a mask prompt $p_m$. Internally, the model applies DDIM inversion to $\mathbf{x}_0$ to obtain its corresponding initial latent $\mathbf{z}^*_T$ and then performs the generation process starting from $\mathbf{z}^*_T$ while incorporating $p$ to generate the target image, $\hat{\mathbf{x}}_0$, applying P2P \cite{hertz2022prompt} as one of the internal components. A key distinction of LAMS-Edit from existing methods is its use of the sequence of latent variables and attention maps computed during the DDIM inversion process, denoted as $\mathbf{z}^*$ and $\mathbf{A}^*$ in Fig.~\ref{fig:lamsedit_overview_real_img}. Algorithm \ref{alg:lamsedit_existing_images} provides pseudo code of LAMS-Edit without masking or style transfer. 

\begin{algorithm}[t]
\caption{LAMS-Edit (base algorithm)}
\label{alg:lamsedit_existing_images}
\LinesNumbered
\SetAlgoLined

\KwIn{An input image $\mathbf{x}_0$, an original prompt $p_o$, a target prompt $p$, and scheduler parameters $(s^{\mathbf{A}}, s^{\mathbf{z}})$.}
\KwOut{An edited image $\mathbf{\hat{x}}_0$.}

$\{w^{\mathbf{A}}_t\}_{t=1}^{T} \gets \text{Scheduler}(s^{\mathbf{A}})$\;
$\{w^{\mathbf{z}}_t\}_{t=0}^{T-1} \gets \text{Scheduler}(s^{\mathbf{z}})$\;


$\mathbf{z}^*_0 \gets \mathcal{E}(\mathbf{x}_0)$\;
$\{\mathbf{z}^*_t\}_{t=1}^{T}, \{\mathbf{A}^*_t\}_{t=1}^{T} \gets \text{InvertDDIM}(\mathbf{z}^*_0, p_o)$\;


$\mathbf{\Tilde{z}}_T \gets \mathbf{z}^*_T$\;
$\mathbf{\hat{z}}_T \gets \mathbf{z}^*_T$\;
\For{$t \gets T$ to $1$}{
    $\Tilde{\mathbf{z}}_{t-1},\Tilde{\mathbf{A}}_{t} \gets \text{DM}(\Tilde{\mathbf{z}}_{t},p_o)$\;
    $\hat{\mathbf{A}}_t\leftarrow \text{DM}
    (\hat{\mathbf{z}}_t, p)$\;
    $\hat{\mathbf{A}}^\text{mixed}_t\gets w^\mathbf{A}_t\cdot\mathbf{A}^*_t+(1-w^\mathbf{A}_t)\cdot\hat{\mathbf{A}}_t$\;
    $\hat{\mathbf{z}}_{t-1}\leftarrow \text{DM}
    (\hat{\mathbf{z}}_t, p)\{\hat{\mathbf{A}}_t\leftarrow \text{P2P}(\tilde{\mathbf{A}}_t,\hat{\mathbf{A}}_t^\text{mixed})\}$\;
    $\mathbf{\hat{z}}^{\text{mixed}}_{t-1} \gets w^\mathbf{z}_{t-1} \cdot \mathbf{z}^*_{t-1} + (1 - w^\mathbf{z}_{t-1}) \cdot \mathbf{\hat{z}_{t-1}}$\;
    $\hat{\mathbf{z}}_{t-1}\gets \mathbf{\hat{z}}^{\text{mixed}}_{t-1}$\;
}
$\hat{\mathbf{x}}_0 \gets \mathcal{D}(\hat{\mathbf{z}}_0)$\;
\Return $\hat{\mathbf{x}}_0$\;
\end{algorithm}

\subsection{P2P Applied to Real Image Editing}

Before explaining the proposed method in detail, we first summarize the computations of P2P when applied to edit a real image.

To edit a real image $\mathbf{x}_0$, we first need to obtain its initial latent variable. Specifically, the image $\mathbf{x}_0$ is mapped by an encoder $\mathcal{E}$ into the latent variable space, producing $\mathbf{z}_0^*$. Then, DDIM inversion is applied to obtain the initial latent variable $\mathbf{z}_T^*$ corresponding to $\mathbf{x}_0$. Notably, this process requires an original prompt $p_o$ that describes $\mathbf{x}_0$. As will be explained later, our method leverages the latent variables and attention maps extracted at intermediate steps of this inversion process, denoted as $\{\mathbf{z}_t^*\}_{t=1}^T$ and $\{\mathbf{A}_t^*\}_{t=1}^T$\footnote{In the case of editing a generated image, $\{\tilde{z}_t\}_{t=1}^T$ and $\{\tilde{A}_t\}_{t=1}^T$ defined below will be used as substitutes for $\{z_t^*\}_{t=1}^T$ and $\{A_t^*\}_{t=1}^T$.}.

P2P generates an edited image $\hat{\mathbf{x}}_0$ as follows. First, a generation process is carried out, starting from $\mathbf{z}_T^*$ and using the original prompt $p_o$. This process effectively reconstructs the original image $\mathbf{x}_0$ and yields a sequence of latent variables and attention maps $\{(\tilde{\mathbf{z}}_t, \tilde{\mathbf{A}}_t)\}_{t=1}^T$\footnote{Instead of computing all steps at once, computations can be performed at each step of the generation process as below.}, where $\tilde{\mathbf{z}}_T = \mathbf{z}_T^*$.

To generate the target image, another generation process is conducted, similarly starting from $\mathbf{z}_T^*$, but with modifications to the attention maps based on the target prompt $p$. Letting $\hat{\mathbf{z}}_t$ denote the latent variable at step $t$ in this process, the modified attention map is obtained in two steps. First, a partial generation step is executed using $\hat{\mathbf{z}}_t$ and the target prompt $p$ to compute an initial attention map $\hat{\mathbf{A}}_t$:
\begin{equation}
    \hat{\mathbf{A}}_t \leftarrow \text{DM}(\hat{\mathbf{z}}_t,p),
    \label{eqn:attn_computation}
\end{equation}
where $\text{DM}$ represents the single-step denoising computation. The resulting attention map is then merged with $\tilde{\mathbf{A}}_t$ using a function, denoted as $\text{P2P}(\tilde{\mathbf{A}}_t, \hat{\mathbf{A}}_t)$, which is selected based on the editing objective (e.g., word swap, phrase addition, etc.).

Finally, to obtain the updated latent variable, the remaining part of the generation step is performed while replacing the attention map with the newly computed one:
\begin{equation}
    \hat{\mathbf{z}}_{t-1}\leftarrow \text{DM}(\hat{\mathbf{z}}_t, p) \{\hat{\mathbf{A}}_t\leftarrow \text{P2P}(\tilde{\mathbf{A}}_t,\hat{\mathbf{A}}_t)\},
    \label{eqn:z_update}
\end{equation}
where $\{\cdot\leftarrow\cdot\}$ indicates that the attention map is replaced. 

\subsection{Latent and Attention Mixing}
\label{sec:lam}

While DDIM inversion enables real image editing as described above, applying P2P to the generation process starting from the inverted latent often yields suboptimal results due to inversion inaccuracies.

To address this, we propose guiding the generation process by mixing the intermediate latent representations and attention maps extracted from DDIM inversion, $\{(\mathbf{z}_t^*, \mathbf{A}_t^*)\}_{t=1}^T$, with those corresponding to the edited images, $\{(\hat{\mathbf{z}}_t, \hat{\mathbf{A}}_t)\}_{t=1}^T$. The goal is to better align the generation path of the edited image with the inversion path, thereby improving structure preservation throughout the process. This approach is motivated by previous studies demonstrating that intermediate latent representations and attention maps encode critical spatial information about the generated image \cite{avrahami2023blendedlatent, mao2023Guided, hertz2022prompt, liu2024freepromptedit}.

The mixing of attention maps and latent variables is performed similarly but independently and at different timings. Details are provided below.

\medskip
\noindent
\textbf{Attention Mixing.}~~ The attention maps are mixed as follows. We apply a weighted linear interpolation between the inverted attention map, $\mathbf{A}_t^*$, and the edited attention map, $\hat{\mathbf{A}}_t$, as:
\begin{equation}
\mathbf{\hat{A}}^{\text{mixed}}_t = w^\mathbf{A} \cdot \mathbf{A}^*_t + (1 - w^\mathbf{A}) \cdot \hat{\mathbf{A}}_t,
\label{eq:lams_attention_mixing}
\end{equation}
where $w^\mathbf{A}\in[0,1]$ is a controllable scale parameter. This mixing is performed after computing $\hat{\mathbf{A}}_t$ using (\ref{eqn:attn_computation}) and the resulting $\mathbf{\hat{A}}^{\text{mixed}}_t$ is used for P2P as $\text{P2P}(\tilde{\mathbf{A}}_t,\mathbf{\hat{A}}^{\text{mixed}}_t)$. We expect this approach to effectively guide the denoising process. It is shown that attention maps play a crucial role in preserving the coarse-grained structure of the original image and maintaining semantic alignment in diffusion models \cite{feng2023structureddiff, hertz2022prompt, liu2024freepromptedit}.

\medskip
\noindent
\textbf{Latent Mixing.}~~ We employ a similar mechanism to mix the latent representations. This is performed after (\ref{eqn:z_update}): once $\hat{\mathbf{z}}_{t-1}$ is obtained using the mixed attention maps and the target prompt $p$, it is then merged with the inversion-derived latent $\mathbf{z}^*_{t-1}$ as follows:
\begin{equation}
\mathbf{\hat{z}}^{\text{mixed}}_{t-1} = w^\mathbf{z} \cdot \mathbf{z}^*_{t-1} + (1 - w^\mathbf{z}) \cdot \mathbf{\hat{z}_{t-1}},
\label{eq:lams_latent_mixing}
\end{equation}
where $w^{\mathbf{z}} \in [0, 1]$ is a controllable scale parameter. We anticipate that this method, particularly when applied to latent variables at earlier steps $(t \sim T)$, will help reinforce the structural information of the original image. This is based on the observation that low-frequency content forms in early steps and high-frequency details in later steps, with these being encoded in the intermediate latent representations \cite{rombach2021stablediffusion, lee2024betasampling, qian2024boosting, mao2023Guided}.

\begin{figure*}[t]
    \centering
    \includegraphics[width=0.9\linewidth]{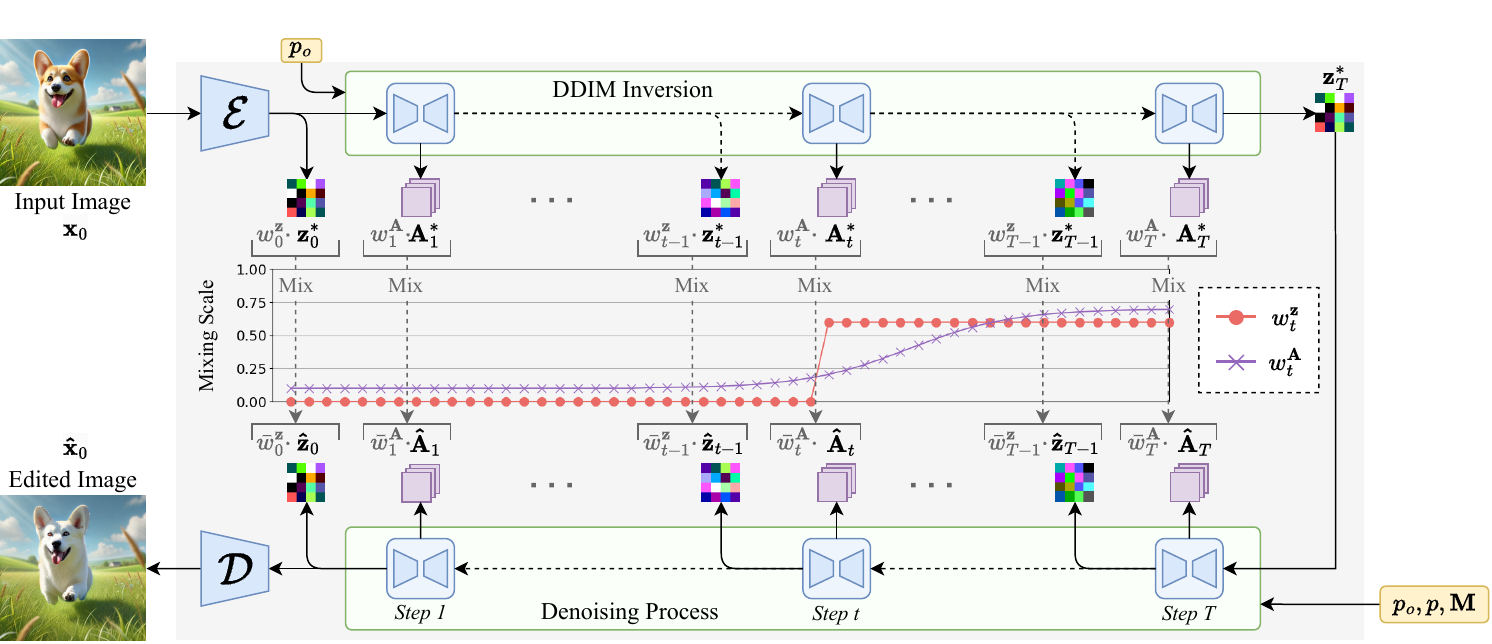}
    \captionof{figure}{Overview of LAMS. At each inversion step, the latent and attention maps are extracted and mixed with their counterparts in the generation process using independent schedulers. The mixing procedures are computed as shown in Eq.~\eqref{eq:lams_latent_mixing} and ~\eqref{eq:lams_attention_mixing}. For clarity in the diagram, we denote $\bar{w}^\mathbf{z}_t = 1-w^\mathbf{z}_t$ and $\bar{w}^\mathbf{A}_t = 1-w^\mathbf{A}_t$, and omit P2P and LoRA for simplicity.}
    \label{fig:lams_schedulers}
\end{figure*}

\subsection{Scheduling Mixing Weights}
\label{sec:lams_schedulers}

As described above, our method mixes the latent variables and attention maps obtained from DDIM inversion with those from the generation process of the edited image. In general, there is often a conflict between preserving the original image's structure and faithfully adhering to the user's editing specifications. To improve the trade-off between these two aspects as much as possible, we introduce schedulers that adjust the mixing rates, $w^\mathbf{z}$ and $w^\mathbf{A}$, at different diffusion steps $t = T, \ldots, 1$.

Through several preliminary experiments, we found that a decaying scheduling pattern—where a higher proportion of inversion-derived representations is used in the early denoising steps and gradually reduced in the later steps—yields the best results. This finding is consistent with previous studies, which have shown that the early steps of the generation process primarily establish the overall structure of an image, while later steps refine fine details \cite{lee2024betasampling, castillo2023adaptive, qian2024boosting, wang2024likepainters}. We also found that it is beneficial to use separate schedulers for latent mixing and attention mixing for the best results. However, the balance between preserving the image structure and adhering to the user's edit request ultimately depends on the user's preference. 

Based on these insights, we designed schedulers with the following four control parameters. By adjusting these parameters, users can customize the mixing process to some extent.
\begin{itemize}
    \item \textbf{Starting scale} $s_{\text{start}} \in [0, 1]$: Proportion of the inverted representations used at the start of denoising.
    \item \textbf{Ending scale} $s_{\text{end}} \in [0, 1]$: Final proportion after decay.
    \item \textbf{Decay until step} $s_{\text{until}} \in [1, T]$: Denoising step by which the scale decays to $s_{\text{end}}$.
    \item \textbf{Decay function type} $s_{\text{type}}$: Controls the decay pattern, with options for stepped, linear, negative exponential, and logistic decay (see supplementary materials for details).
\end{itemize}
Figure \ref{fig:lams_schedulers} illustrates the internal mechanism of LAMS-Edit, highlighting the role of the schedulers. The pseudo code for LAMS-Edit, incorporating the schedulers, is provided in Algorithm~\ref{alg:lamsedit_existing_images}.

\subsection{SAM-Guided Masking}
\label{sec:lamsedit_sam_masking}

LAMS-Edit can be used with latent masking \cite{avrahami2023blendedlatent} to isolate specific regions for modification, further improving the spatial accuracy of edits. Existing methods using this technique \cite{cao_2023_masactrl,couairon2023diffedit,park2024shape,simsar2023lime,wang2023instructedit} generally use either internal representations from diffusion models, such as attention maps, or external models for mask generation. We adopt the latter approach by using the Segment Anything Model (SAM) \cite{kirillov2023sam} to generate a binary region of interest (ROI) mask $\mathbf{M}$ for the input image, which is resized to match the dimensions of the latent representation $\mathbf{\hat{z}}_t$. To apply the mask, the update of $\hat{\mathbf{z}}_{t-1}$ at the final step of each generation process (i.e., line 12 of Algorithm \ref{alg:lamsedit_existing_images}) is performed as follows:
\begin{equation}
\mathbf{\hat{z}}_{t-1} \leftarrow \mathbf{M} \odot \mathbf{\hat{z}}^{\text{mixed}}_{t-1} + (1 - \mathbf{M}) \odot \mathbf{z}^*_{t-1},
\label{eq:masking}
\end{equation}
where $\odot$ denotes the element-wise multiplication.

\subsection{Style Transfer with LoRA}
\label{sec:lamsedit_style_transfer}

In LAMS-Edit, LoRA \cite{hu2022lora} can also be incorporated into the diffusion model, enabling style transfer either independently or simultaneously with editing.
This is achieved by simply loading the LoRA checkpoint after DDIM inversion and before initiating the generation process (i.e., between lines 4 and 5 in Algorithm~\ref{alg:lamsedit_existing_images}). Since LoRA and LAMS function independently, style transfer can be seamlessly applied alongside text-guided editing. The complete algorithm is provided in the supplementary material. See Sec.~\ref{sec:qualitative_eval} for experimental results and discussions.

\section{Experiments}

We conducted a series of experiments to evaluate our method. 
For the diffusion model, we use Stable-Diffusion-v1-5 \cite{sd15} for photorealistic images and Anything-V4 \cite{anything4} for anime-style images. Automatic mask generation is performed using the Panoptic SAM implementation \cite{panopticsam2024}, based on SAM \cite{kirillov2023sam}, with a text-aware pipeline applied to segmentation tasks. Unless otherwise specified, all experiments follow the settings of prior studies, using 50 steps for both DDIM inversion and generation, with a guidance scale set to 7.5 \cite{Tumanyan_2023_CVPR,hertz2022prompt,cao_2023_masactrl}.

We evaluate our method under the above configuration by comparing it with state-of-the-art (SOTA) approaches, including DiffEdit \cite{couairon2023diffedit}, Pix2Pix-Zero \cite{parmar2023pix2pixzero}, SDEdit \cite{meng2022sdedit}, Plug-and-Play (PnP) \cite{Tumanyan_2023_CVPR}, LEDITS++ \cite{brack2023leditspp}, PnPInversion (PnPInv) \cite{ju2023pnpinv} and Null-text Inversion (NTI) \cite{mokady2022null}, combined with P2P for image editing as proposed in its original work. Unless otherwise specified, all methods use their default hyperparameters. The parameters of our scheduler are fixed to the default values provided in the supplementary materials.

Some methods are compatible with the inclusion of an additional mask input. To ensure a fair comparison, we use the same SAM-generated mask for all methods that allow masking. In the following, we compare all methods, including our own, in two groups: with and without mask input. Hereafter, `Ours' refers to our method without SAM-guided masking, while `Ours (w/ mask)' includes masking.

\subsection{Quantitative Evaluation}

First, we present the results of the quantitative comparison. Due to the lack of standard datasets, we constructed a dataset of 100 randomly sampled images from COCO2017 \cite{lin2015coco}, covering a diverse range of objects suitable for various editing tasks. For each image, we use the corresponding caption provided by the dataset as the original prompt, while the target prompt was manually created to test different editing scenarios.

\begin{figure}[t]
    \centering
    \includegraphics[width=\linewidth]{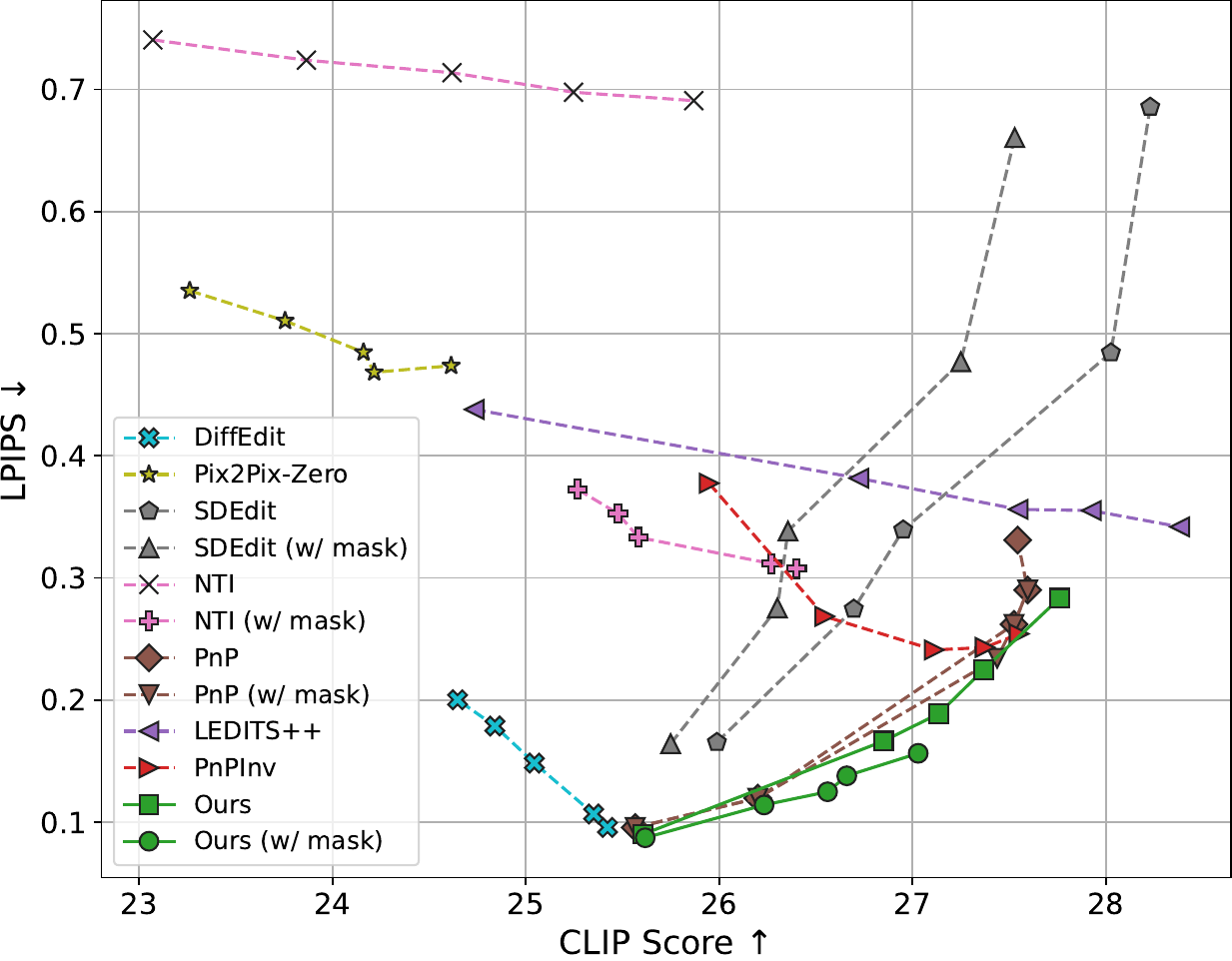}
    \captionof{figure}{Fidelity-editability trade-off of the image editing methods. Closer proximity to the lower right indicates a better balance.}
    \label{fig:lpips_vs_clipscore}
\end{figure}

One of the major challenges in image editing and generation is the {\em fidelity-editability trade-off}. This refers to the inherent conflict between preserving the original content (fidelity) and applying edits as intended (editability), making it difficult to achieve both simultaneously \cite{couairon2023diffedit, zou2024instdiffedit, kawar2023imagic}. 
To assess the extent of this trade-off in different methods, we employ two widely used metrics in the image generation and editing domain: LPIPS (lower is better) for content preservation and CLIP Score (higher is better) for alignment with intended edits. 
Figure \ref{fig:lpips_vs_clipscore} presents the trade-off curves for the compared methods, with data points obtained by varying the starting timestep of the generation process. Methods positioned toward the lower right of the graph—indicating both lower LPIPS and higher CLIP Score—are considered superior. As shown, existing methods—including DiffEdit, Pix2Pix-Zero, NTI, and LEDITS++—achieve only suboptimal trade-offs.
This quantitative evaluation aligns with the observed tendency of these methods to introduce artifacts or distort content, or to fall short in producing meaningful edits.
In contrast, our method, particularly ours (w/ mask), achieves the best trade-off, effectively balancing fidelity and editability.

\subsection{Qualitative Evaluation}
\label{sec:qualitative_eval}

We also conducted a qualitative comparison using a diverse set of images, including natural and synthetic images generated by DALL·E 3 and Stable Diffusion.

\begin{figure*}[t]
    \centering
    \includegraphics[width=\linewidth]{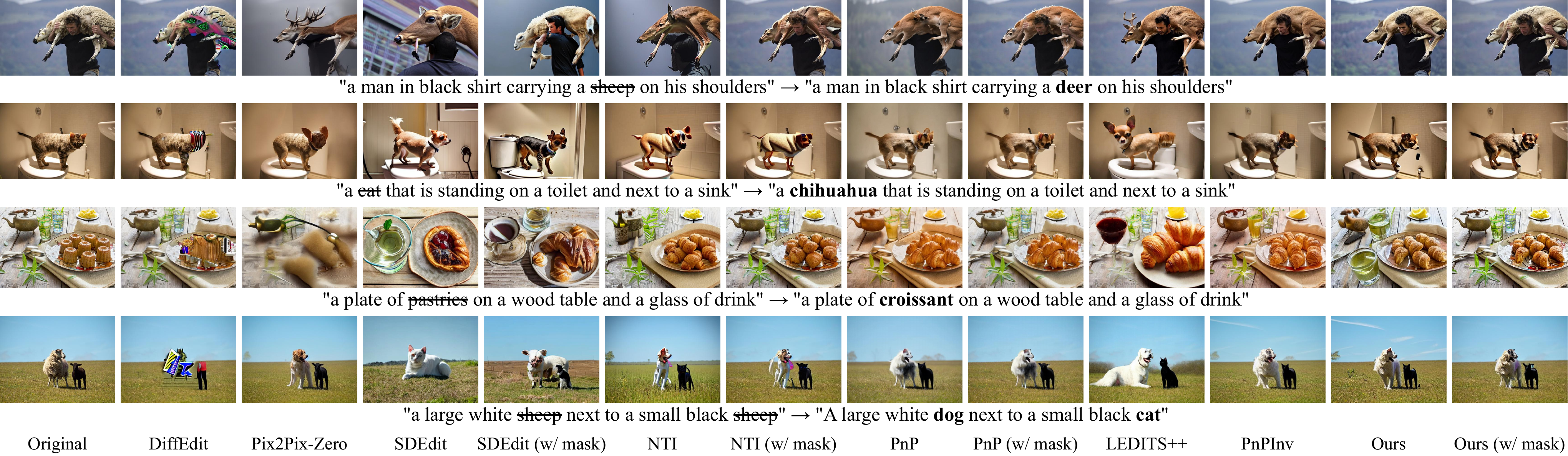}
    \captionof{figure}{Results of different image editing methods for different editing scenarios.}
    \label{fig:image_editing_comparison}
\end{figure*}

\para{Image Editing.} 
Figure \ref{fig:image_editing_comparison} shows a qualitative comparison of the methods with several examples. It demonstrates that our method outperforms baselines by achieving semantically accurate edits while preserving the original content. Among methods without extra mask input, others struggle with convincing edits (e.g., Pix2Pix-Zero in rows 1 to 3), fail to maintain structural integrity (e.g., SDEdit and LEDITS++), or introduce artifacts (e.g., DiffEdit and NTI). In contrast, our method effectively generates edits with the overall structure preserved. For methods using extra mask input, NTI (w/ mask) better preserves non-targeted areas but still introduces artifacts, while SDEdit (w/ mask) improves structural preservation, but not sufficiently. Although PnP and PnPInv produce results similar in quality to our method, our method strikes a better balance between content preservation and meaningful edits, as demonstrated by the fidelity-editability trade-off in Fig.~\ref{fig:lpips_vs_clipscore}.

\begin{figure}[t]
    \centering
    \includegraphics[width=\linewidth]{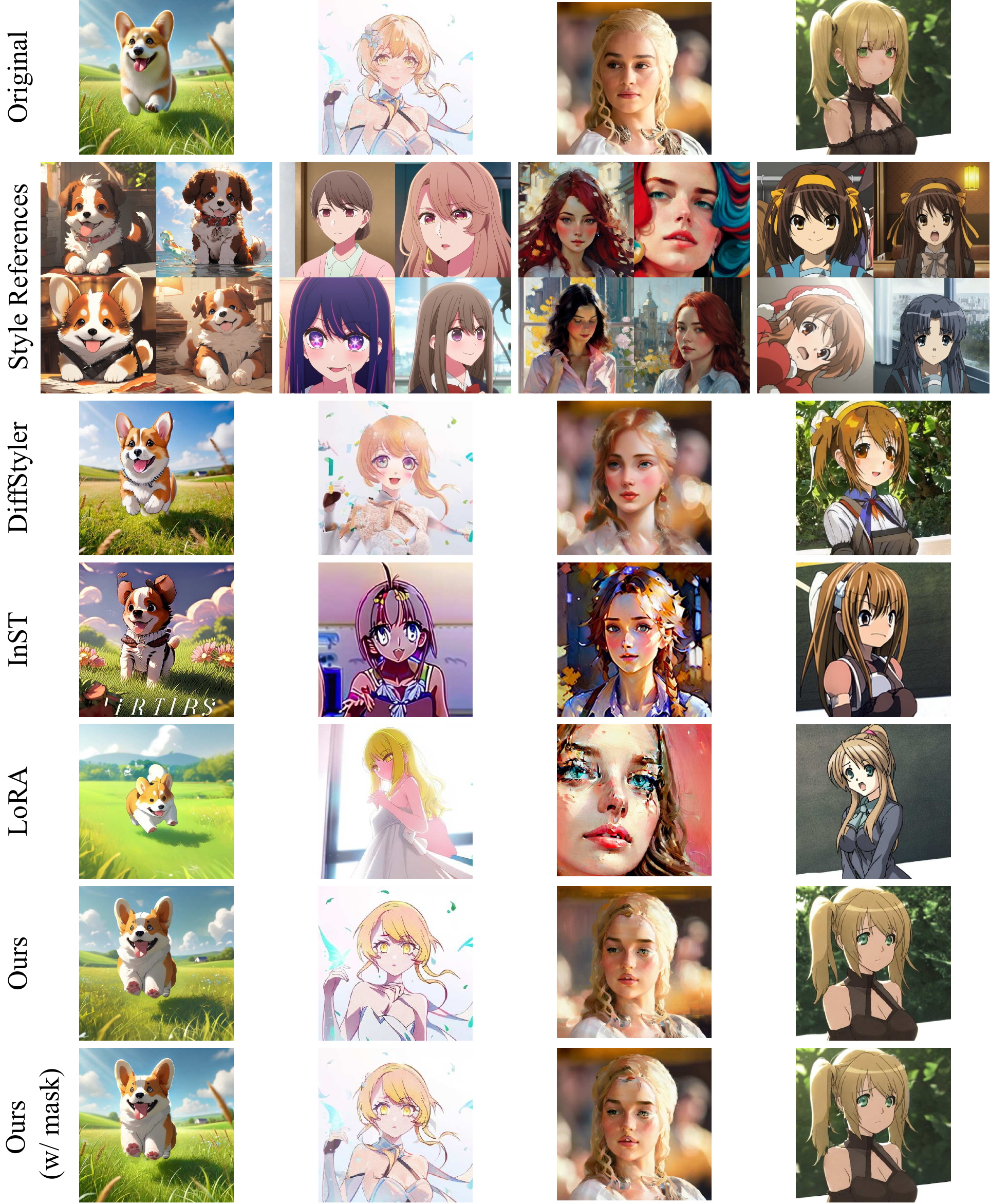}
    \captionof{figure}{Qualitative comparison of 
    style transfer.} 
    \label{fig:style_transfer_comparison}
\end{figure}

\para{Style Transfer.} As described in Sec.~\ref{sec:lamsedit_style_transfer}, LAMS can be combined with a LoRA-based style transfer method, enabling simultaneous content preservation and style transformation. We compare our method with DiffStyler \cite{li2024diffstyler}, InST \cite{Zhang_2023_inst}, and a simple LoRA-based baseline (hereafter referred to as ``LoRA'') that performs DDIM inversion, loads LoRA, and then runs the reverse diffusion process. Fig.~\ref{fig:style_transfer_comparison} shows several examples. While LoRA adapts styles, it struggles to preserve content; DiffStyler maintains content well but compromises on identity retention; and InST applies styles effectively yet often distorts character identities. In contrast, our method preserves both content and identity while faithfully incorporating styles, with a mask (specified via the prompt) further protecting key elements and enhancing content integrity.
Since style transfer is hard to evaluate quantitatively, we conducted a user study comparing five approaches—including our method with and without masking. For 15 images each subjected to a different style transfer, 41 participants were shown both the original and the transformed images and asked to vote on which method was superior in terms of content preservation, style application, and overall quality. The results, presented in Table~\ref{tab:user_study_results}, indicate that our method (with mask) outperformed all baselines, demonstrating the effectiveness and robustness of LAMS-Edit in style transfer tasks.

\begin{table}
\footnotesize
\centering
\begin{tabular}{p{0.7cm} p{1cm} p{0.7cm} p{0.7cm} p{0.7cm} p{1.75cm}}
\hline
& DiffStyler & InST & LoRA & Ours & Ours (w/~mask) \\
\hline
Content & 16.6\% & 2.9\% & 1.5\% & 15.1\% & 63.9\% \\
Style & 18.9\% & 19.7\% & 8.8\% & 8.5\% & 44.2\% \\
Overall & 27.8\% & 6.8\% & 2.6\% & 11.9\% & 50.9\% \\
\hline
\end{tabular}
\caption{Human evaluation of style transfer methods in terms of content preservation, style application, and overall quality.}
\label{tab:user_study_results}
\end{table}

\para{Image Editing with Style Transfer.} Our method enables the simultaneous application of image editing and style transfer, successfully balancing content modification and style adaptation as intended. Several examples are shown in Fig.~\ref{fig:teaser}. This highlights the practicality of our method for real-world applications where both content and style modifications are required.

\subsection{Alabtion Study}

\begin{figure}[t]
    \centering
    \includegraphics[width=\linewidth]{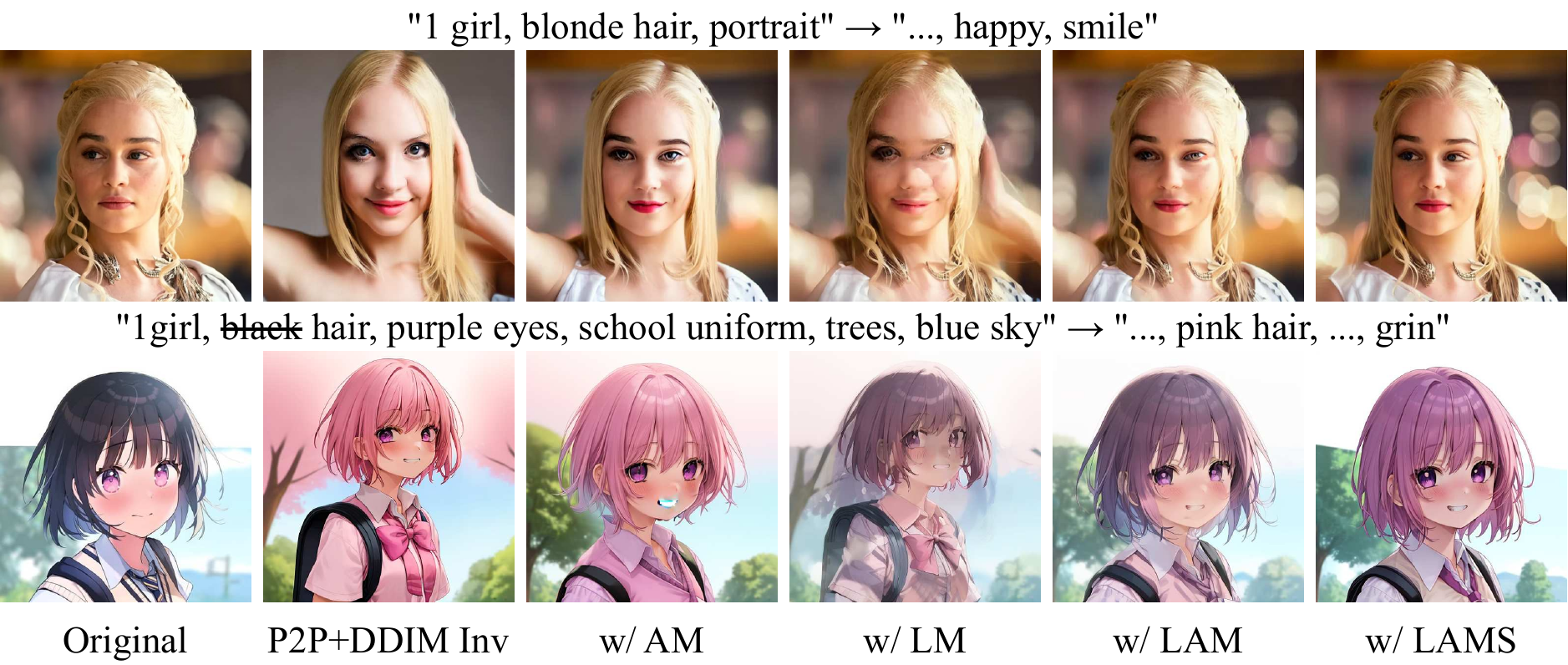}
    \captionof{figure}{Ablation study of the LAMS components for image editing task. }
    \label{fig:ablation_lams}
\end{figure}

\para{Effectiveness of LAMS.} We conduct an ablation study on LAMS by evaluating its components—Attention Mixing (AM), Latent Mixing (LM), their combination (LAM), and the full pipeline with Schedulers (LAMS). These are compared against the baseline ``P2P+DDIM Inv,'' which integrates P2P with DDIM inversion. Examples are shown in Fig.~\ref{fig:ablation_lams}. While ``P2P+DDIM Inv'' adapts styles effectively, it struggles with content preservation. AM improves structural integrity but fails to retain individual identity (first row), while LM introduces layering artifacts. Combining AM and LM (LAM) enhances fidelity and semantic accuracy but still leaves some artifacts. Adding schedulers (LAMS) to adjust the mixing scales preserves individual identity (first row) and achieves a better balance between content preservation and meaningful edits. 

\begin{figure}[t]
    \centering
    \includegraphics[width=\linewidth]{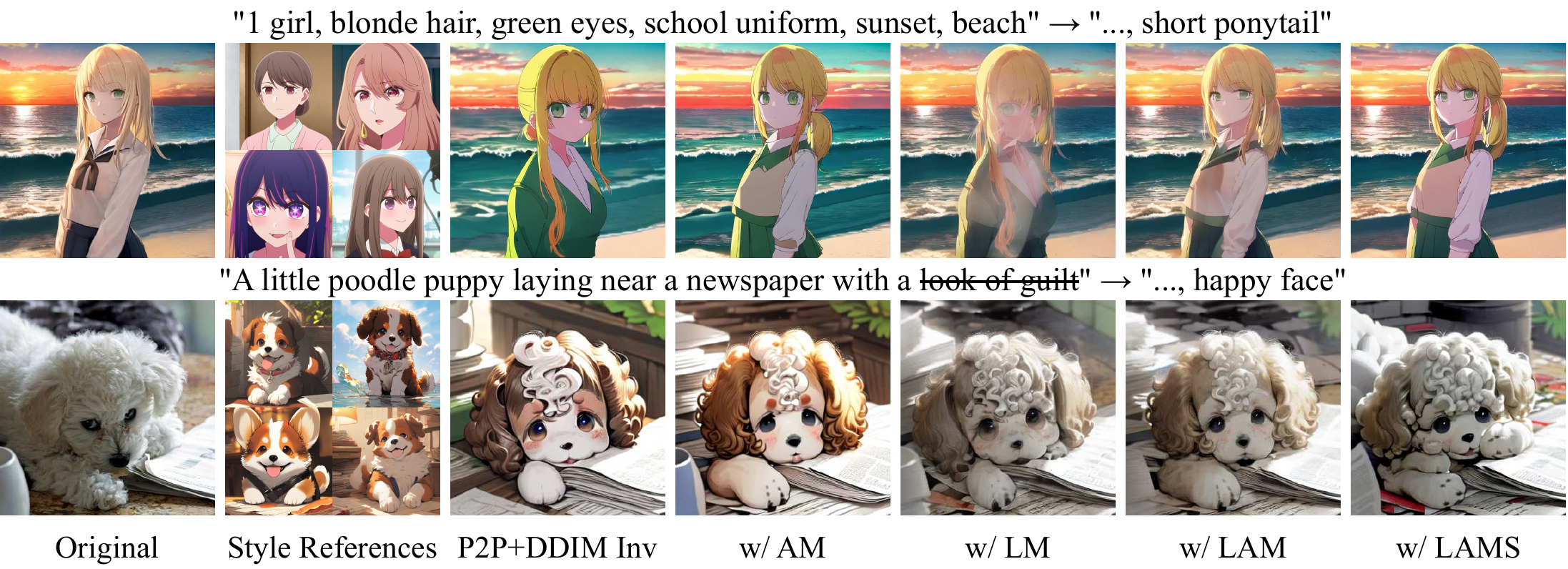}
    \captionof{figure}{Ablation study of the LAMS components for image editing combined with style transfer tasks.}
    \label{fig:ablation_lams_2}
\end{figure}

Similarly, Fig.~\ref{fig:ablation_lams_2} presents the ablation results for style transfer tasks. The baseline ``P2P+DDIM Inv'' effectively transfers style but struggles to preserve content. Integrating AM slightly enhances structure preservation, while LM improves detail retention but introduces layering effects. Combining both (LAM) achieves a better balance between content preservation and style application. Finally, incorporating the full LAMS framework further strengthens content preservation while maintaining effective style transfer.

The ablation study shows that latent mixing enhances fine details and pixel-level content preservation, while attention mixing contributes to better semantic and structural consistency. Combining these with Schedulers further promotes a smoother balance between content preservation and desired edits.

\section{Conclusion}
In this paper, we have presented LAMS-Edit, a unified framework for text-to-image editing and style transfer. At its core is LAMS, a novel method that enhances structural preservation by guiding the denoising trajectory through the scheduled mixing of inverted latent and attention maps. LAMS-Edit also integrates SAM-guided masking for precise localized editing. Our approach achieves a superior fidelity-editability trade-off compared to existing methods, advancing image editing and style transfer with a tuning-free, efficient design.
\clearpage

{
    \small
    \bibliographystyle{ieeenat_fullname}
    \bibliography{main}
}

\clearpage
\setcounter{page}{1}
\maketitlesupplementary

\section{Method and Implementation Details}
\label{sup:method_details}

\subsection{Algorithm}
\label{sup:algorithm}

Algorithm~\ref{alg:lamsedit_existing_images_full} outlines the complete process for editing real images, incorporating LoRA for style transfer (lines 5–7) and SAM-guided masking for localized edits (line 16). The LoRA checkpoint is loaded after DDIM inversion and before the reverse diffusion process, ensuring that the inverted representations preserve the original structure while enabling style transformation during reverse diffusion.

\begin{algorithm}[h]
\caption{LAMS-Edit (full algorithm)}
\label{alg:lamsedit_existing_images_full}
\LinesNumbered
\SetAlgoLined

\KwIn{An input image $\mathbf{x}_0$, an original prompt $p_o$, a target prompt $p$, and scheduler parameters $(s^{\mathbf{A}}, s^{\mathbf{z}})$.}
\KwOut{An edited image $\mathbf{\hat{x}}_0$.}

$\{w^{\mathbf{A}}_t\}_{t=1}^{T} \gets \text{Scheduler}(s^{\mathbf{A}})$\;
$\{w^{\mathbf{z}}_t\}_{t=0}^{T-1} \gets \text{Scheduler}(s^{\mathbf{z}})$\;


$\mathbf{z}^*_0 \gets \mathcal{E}(\mathbf{x}_0)$\;
$\{\mathbf{z}^*_t\}_{t=1}^{T}, \{\mathbf{A}^*_t\}_{t=1}^{T} \gets \text{InvertDDIM}(\mathbf{z}^*_0, p_o)$\;

\If{$L$ is provided}{
    $DM \gets \text{LoadLoRA}(DM, L)$\;
}

$\mathbf{\Tilde{z}}_T \gets \mathbf{z}^*_T$\;
$\mathbf{\hat{z}}_T \gets \mathbf{z}^*_T$\;
\For{$t \gets T$ to $1$}{
    $\Tilde{\mathbf{z}}_{t-1},\Tilde{\mathbf{A}}_{t} \gets \text{DM}(\Tilde{\mathbf{z}}_{t},p_o)$\;
    $\hat{\mathbf{A}}_t\leftarrow \text{DM}
    (\hat{\mathbf{z}}_t, p)$\;
    $\hat{\mathbf{A}}^\text{mixed}_t\gets w^\mathbf{A}_t\cdot\mathbf{A}^*_t+(1-w^\mathbf{A}_t)\cdot\hat{\mathbf{A}}_t$\;
    $\hat{\mathbf{z}}_{t-1}\leftarrow \text{DM}
    (\hat{\mathbf{z}}_t, p)\{\hat{\mathbf{A}}_t\leftarrow \text{P2P}(\tilde{\mathbf{A}}_t,\hat{\mathbf{A}}_t^\text{mixed})\}$\;
    $\mathbf{\hat{z}}^{\text{mixed}}_{t-1} \gets w^\mathbf{z}_{t-1} \cdot \mathbf{z}^*_{t-1} + (1 - w^\mathbf{z}_{t-1}) \cdot \mathbf{\hat{z}_{t-1}}$\;
    $\mathbf{\hat{z}}_{t-1} \gets \mathbf{M} \odot \mathbf{\hat{z}}^{\text{mixed}}_{t-1} + (1 - \mathbf{M}) \odot \mathbf{z}^*_{t-1}$\;
}
$\hat{\mathbf{x}}_0 \gets \mathcal{D}(\hat{\mathbf{z}}_0)$\;
\Return $\hat{\mathbf{x}}_0$\;
\end{algorithm}

\subsection{Default Schedulers for Latent and Attention Mixing}
\label{appendix:default_schedulers}

\begin{figure}[t]
    \centering
    \includegraphics[width=\linewidth]{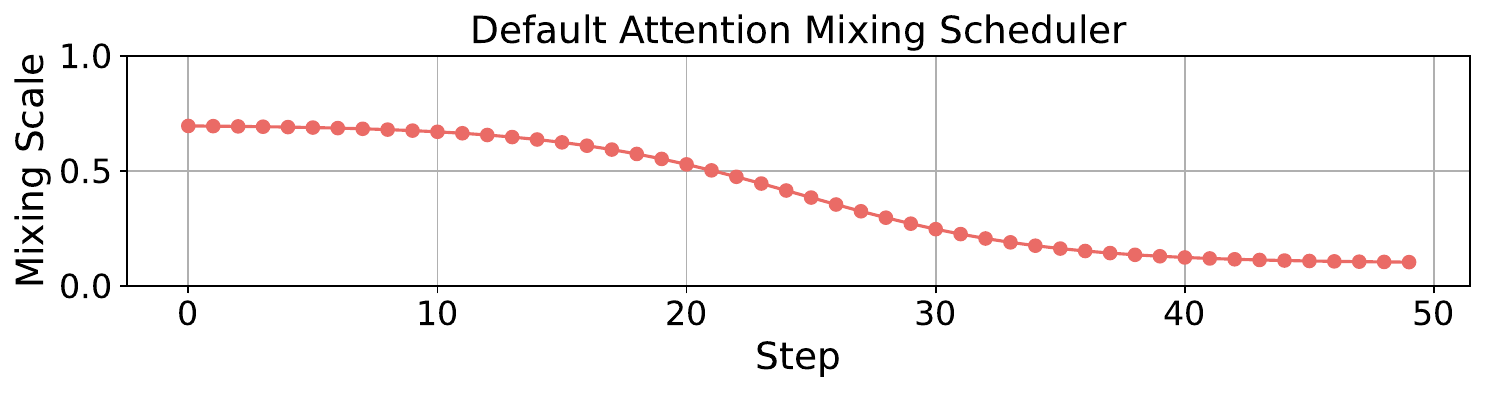}
    \captionof{figure}{Default attention mixing scheduler.}
    \label{fig:default_attention_mixing_scheduler}
\end{figure}

\begin{figure}[t]
    \centering
    \includegraphics[width=\linewidth]{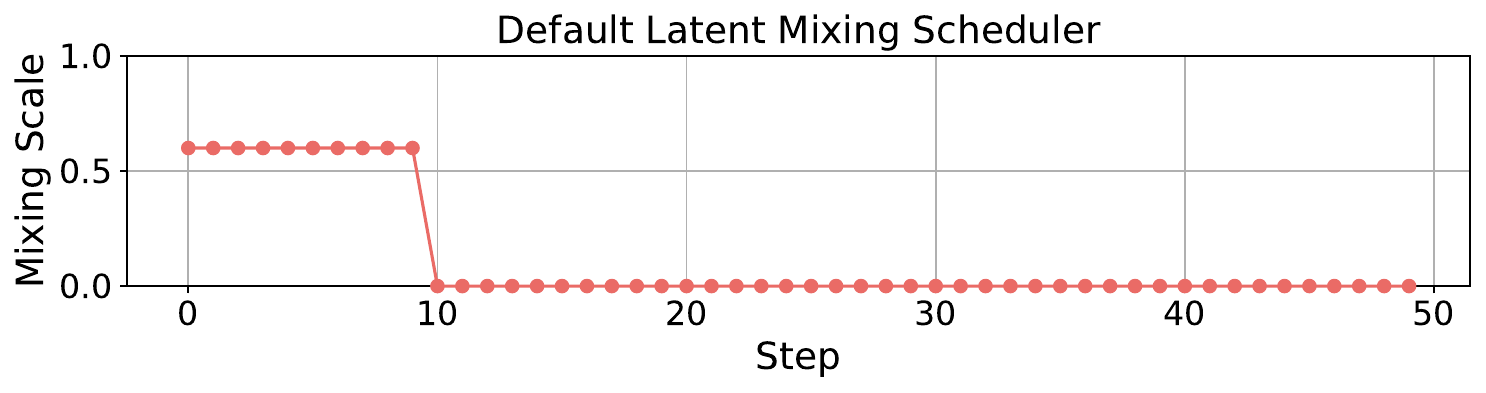}
    \captionof{figure}{Default latent mixing scheduler.}
    \label{fig:default_latent_mixing_scheduler}
\end{figure}

The default parameters for the mixing schedulers used in our experiments were determined empirically (see Sec.~\ref{sup:ablation_study}) and are outlined below:
\begin{itemize}
    \item \textbf{Attention Mixing (\(s^{\mathbf{A}}\))}: \( \text{start}=0.7, \text{end}=0.1, \text{until}=50, \text{type}=\text{logistic} \)
    \item \textbf{Latent Mixing (\(s^{\mathbf{z}}\))}: \( \text{start}=0.6, \text{end}=0.0, \text{until}=10, \text{type}=\text{stepped} \)
\end{itemize}
Figures \ref{fig:default_attention_mixing_scheduler} and \ref{fig:default_latent_mixing_scheduler} illustrate the default schedulers for attention mixing and latent mixing respectively. The precise values for these schedulers are detailed below:
\begin{itemize} 
    \item wA: Default scheduler for attention mixing. 
    \item wz: Default scheduler for latent mixing. 
\end{itemize}

\begin{lstlisting}
wA = [0.696  0.6951 0.694  0.6926 0.691  0.689  0.6866 0.6836 0.68   0.6757 0.6704 0.6641 0.6566 0.6476 0.637  0.6245 0.61   0.5933 0.5742 0.5527 0.5288 0.5028 0.4749 0.4456 0.4153 0.3847 0.3544 0.3251 0.2972 0.2712 0.2473 0.2258 0.2067 0.19   0.1755 0.163  0.1524 0.1434 0.1359 0.1296 0.1243 0.12   0.1164 0.1134 0.111  0.109  0.1074 0.106  0.1049 0.104]

wz = [0.6 0.6 0.6 0.6 0.6 0.6 0.6 0.6 0.6 0.6 0.  0.  0.  0.  0.  0.  0.  0. 0.  0.  0.  0.  0.  0.  0.  0.  0.  0.  0.  0.  0.  0.  0.  0.  0.  0. 0.  0.  0.  0.  0.  0.  0.  0.  0.  0.  0.  0.  0.  0. ]
\end{lstlisting}

\section{Supplementary Results}
\label{sup:supplementary_results}

This section presents additional experimental results that complement the findings discussed in the main text. These supplementary results provide further insights and detailed analyses omitted from the main sections for brevity.

\begin{figure}[t]
    \centering
    \includegraphics[width=\linewidth]{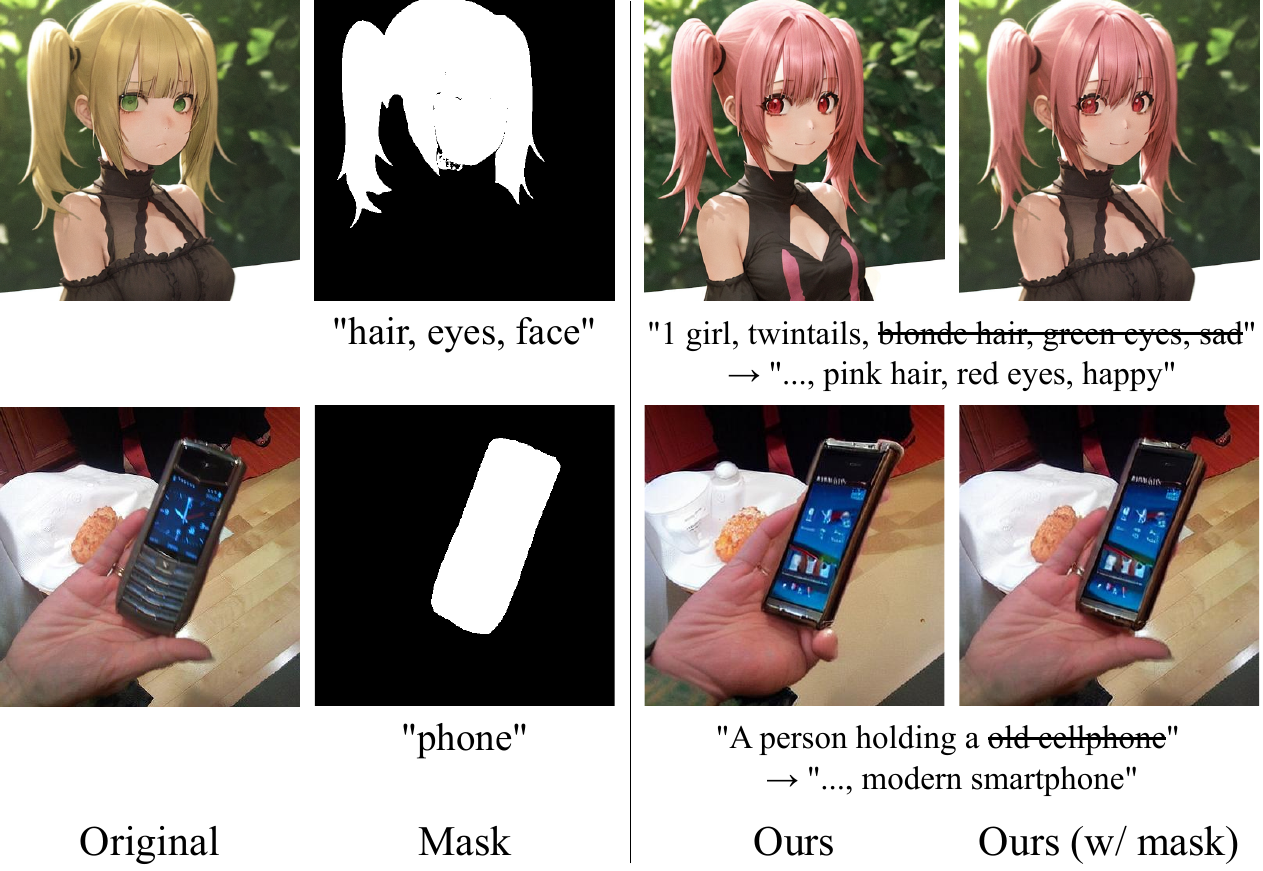}
    \captionof{figure}{Qualitative results of image editing using LAMS-Edit. Our method effectively edits the content while the mask enhances content preservation in non-targeted regions.}
    \label{fig:image_editing_ours}
\end{figure}

\subsection{Image Editing} 
The visual results of our method are showcased in Fig.~\ref{fig:image_editing_ours}. The examples highlight the effectiveness of our method in producing semantically accurate edits while maintaining fidelity to the original content. Notably, Ours (w/ mask) demonstrates improved control over localized edits, ensuring changes are constrained to specific regions defined by the mask. This is especially evident in cases such as modifying a character's hair, an individual's hand, or specific objects, where Ours (w/ mask) better preserves surrounding details compared to Ours.

\begin{figure}[t]
    \centering
    \includegraphics[width=0.9\linewidth]{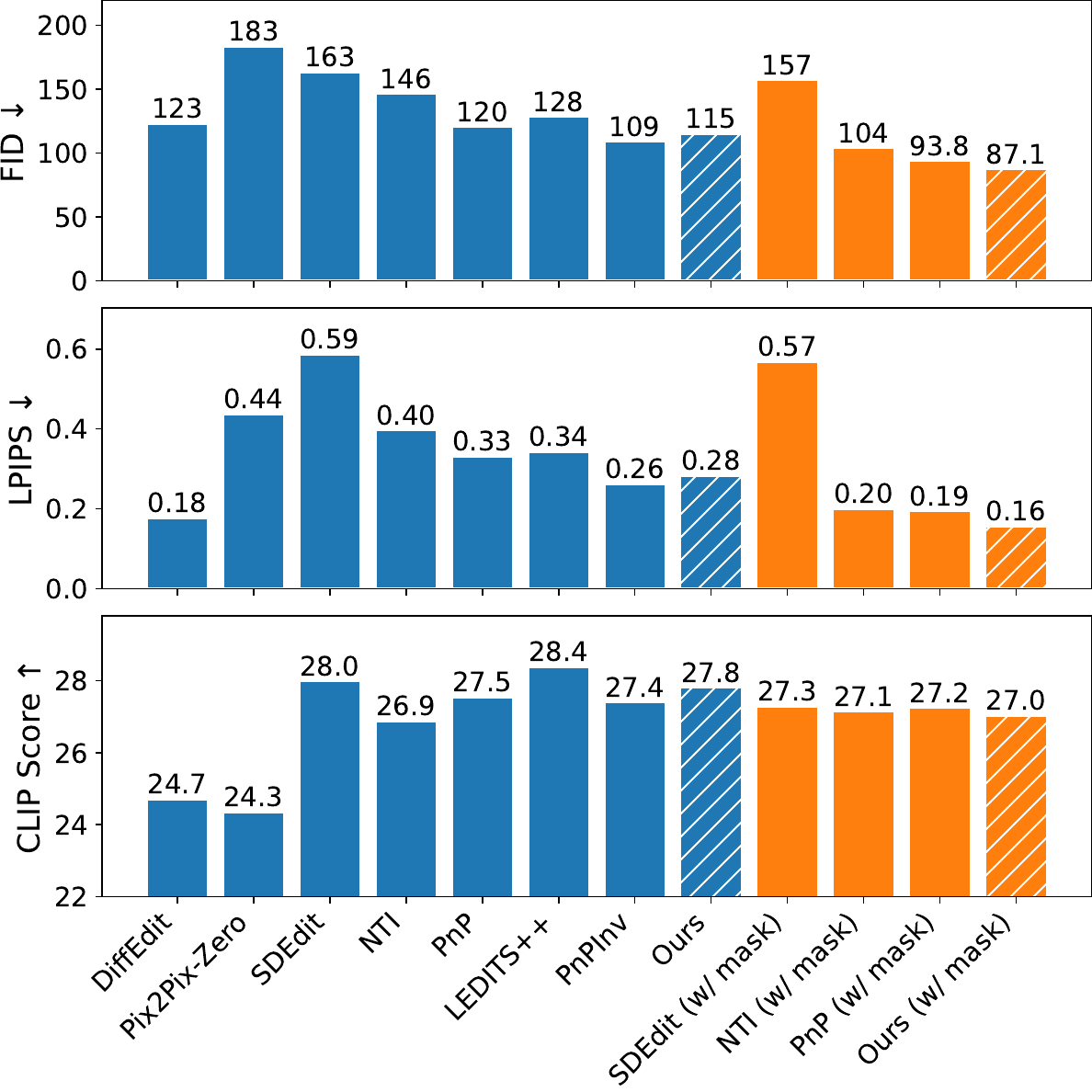}
    \captionof{figure}{
    Quantitative evaluation of the compared image editing methods on our dataset of 100 COCO2017 images using three metrics: FID, LPIPS, and CLIP Score. Methods without masking are shown in blue, while those with masking are shown in orange. See the text for further details. }
    \label{fig:scores_bars}
\end{figure}

To assess the performance of different methods, we report three widely used metrics in the image generation and editing domain: FID, LPIPS, and CLIP Score. FID and LPIPS (lower is better) evaluate fidelity, while CLIP Score (higher is better) measures editability. Figure~\ref{fig:scores_bars} presents the results for the compared methods. Among approaches without mask input, our method achieves relatively low FID and LPIPS scores along with a comparatively high CLIP score. For methods utilizing mask input, our approach achieves a high CLIP score comparable to the others, while obtaining the best FID and LPIPS scores. This demonstrates a superior balance between perceptual fidelity and semantic alignment with the target prompt. Figure~\ref{fig:lpips_vs_clipscore} further illustrates this favorable trade-off achieved by our method.

\begin{figure}[t]
    \centering
    \includegraphics[width=0.9\linewidth]{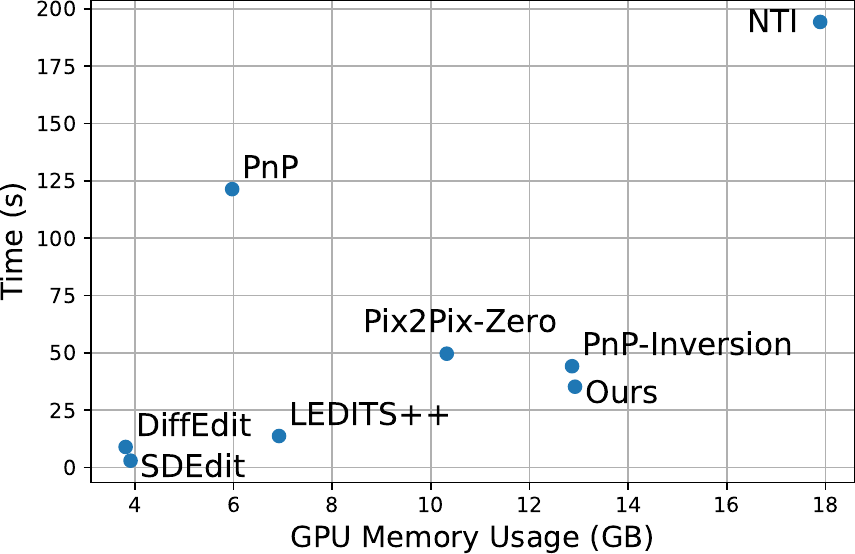}
    \captionof{figure}{Average time and GPU memory usage comparison.}
    \label{fig:computational_costs}
\end{figure}

In addition to evaluating the generated results, Fig.~\ref{fig:computational_costs} presents the runtime and GPU memory consumption for editing a $512\times 512$ image on a TITAN RTX (24GB). Our method also offloads approximately 12GB to CPU memory to store latents and attention maps. While the GPU memory usage is relatively high compared to other optimization-free methods, the runtime remains comparable to the average.

\subsection{Ablation Study}
\label{sup:ablation_study}

\begin{figure}[t]
    \centering
    \includegraphics[width=\linewidth]{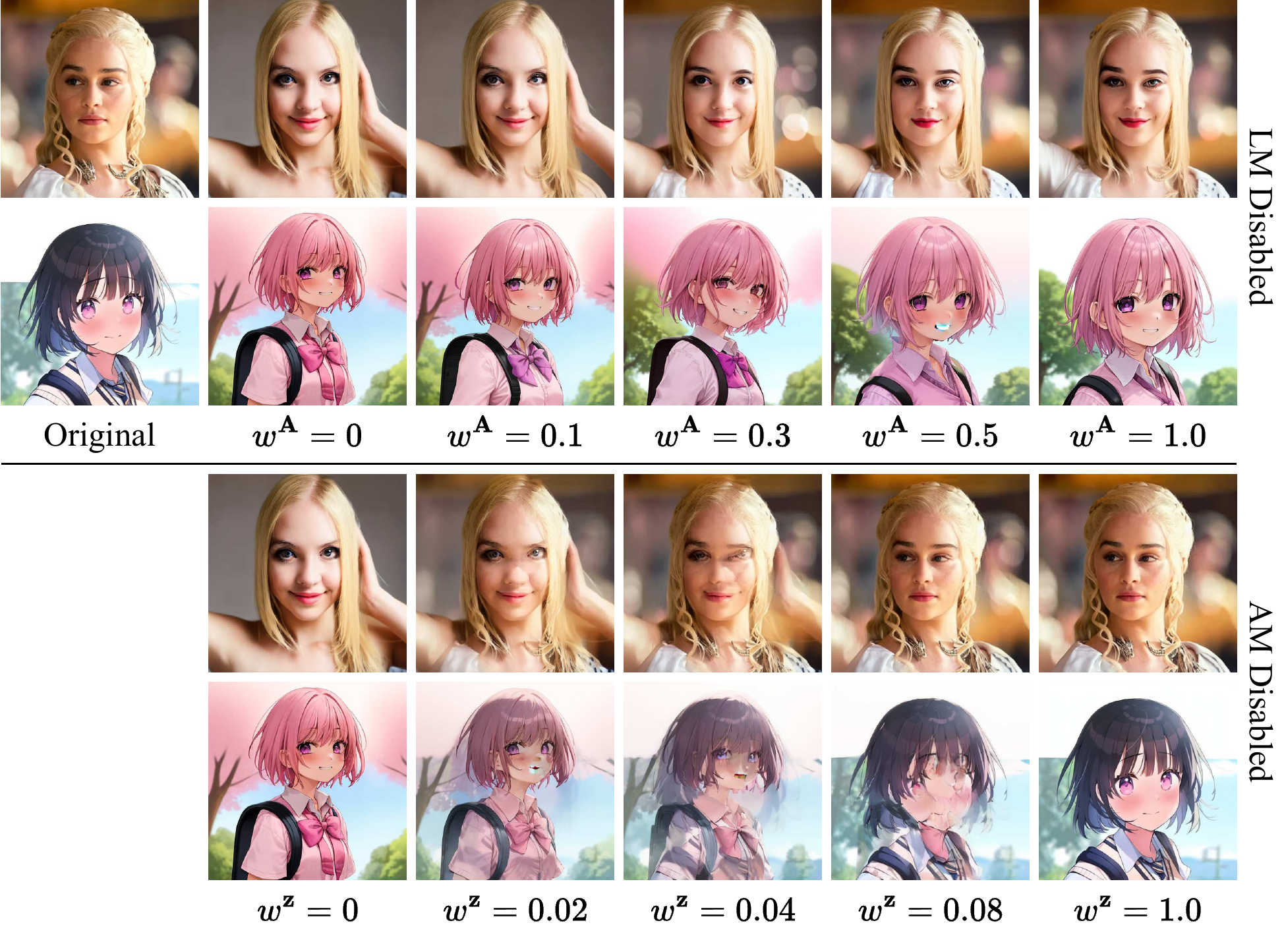}
    \captionof{figure}{The first two rows show the results of varying the attention mixing scale ($w^{\mathbf{A}}$) with latent mixing disabled, while the last two rows show the effects of varying the latent mixing scale ($w^{\mathbf{z}}$) with attention mixing disabled.}
    \label{fig:ablation_am_lm_scale}
\end{figure}

\para{Effect of Varying Mixing Scale.} We also investigate the impact of varying attention and latent mixing scales without schedulers. As shown in Fig.~\ref{fig:ablation_am_lm_scale}, increasing the attention mixing scale enhances structure preservation, with $w^{\mathbf{A}}=1.0$ maintaining the original layout while allowing edits; however, it may alter identity (first row). Increasing the latent mixing scale progressively blends original pixels into the edited image, with $w^{\mathbf{z}}=1.0$ producing an identical reconstruction as it bypasses the diffusion process. These findings show that attention mixing preserves layout and enables semantic edits, while latent mixing retains pixel-level details but reduces editability when applied excessively.

\para{Effect of Mixing Schedulers.} To evaluate the impact of mixing schedulers in LAMS, we adjust the scheduler parameters for attention mixing and latent mixing, denoted as $s^{\mathbf{A}} = (s^{\mathbf{A}}_{\text{start}}, s^{\mathbf{A}}_{\text{end}}, s^{\mathbf{A}}_{\text{until}}, s^{\mathbf{A}}_{\text{type}})$ and $s^{\mathbf{z}} = (s^{\mathbf{z}}_{\text{start}}, s^{\mathbf{z}}_{\text{end}}, s^{\mathbf{z}}_{\text{until}}, s^{\mathbf{z}}_{\text{type}})$, respectively, to identify the most effective scheduling schemes. For these experiments, parameters not being varied, or unless explicitly specified otherwise, will use the default values provided in Appendix~\ref{appendix:default_schedulers}, which were empirically determined.

\begin{figure}[t]
    \centering
    \includegraphics[width=\linewidth]{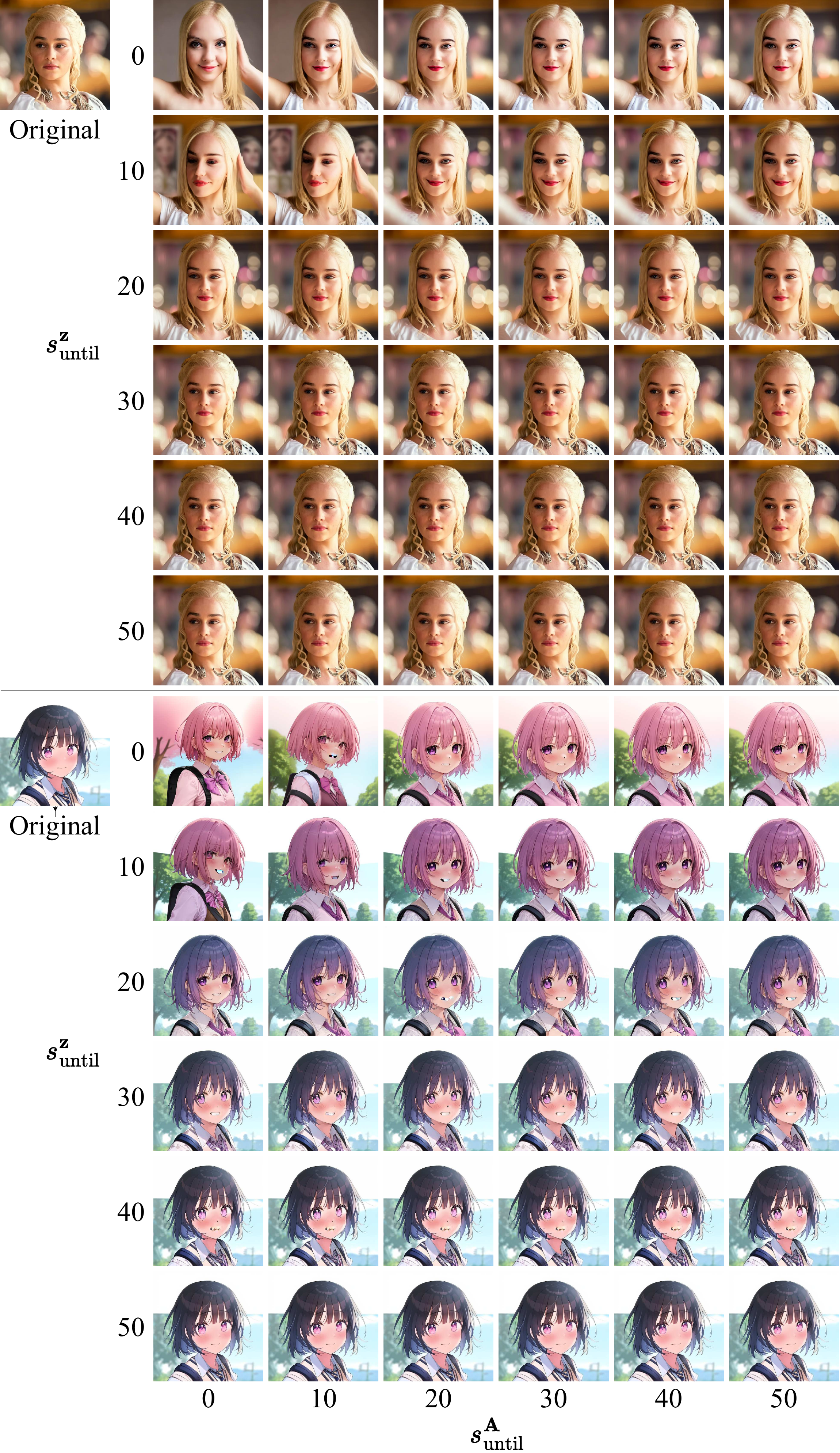}
    \captionof{figure}{Image editing results with varying scheduler parameters for \textbf{Decay until step}: $s^{\mathbf{z}}_{\text{until}}$ for latent mixing and $s^{\mathbf{A}}_{\text{until}}$ for attention mixing.}
    \label{fig:ablation_scheduler_until}
\end{figure}

Figure~\ref{fig:ablation_scheduler_until} compares the effects of varying $s^{\mathbf{z}}_{\text{until}}$ and $s^{\mathbf{A}}_{\text{until}}$, which determine the step at which the mixing scale decays to the target value $s_{\text{end}}$. For this experiment, stepped decay schedulers were used for both operations, as their simplicity makes it easier to observe changes. The results suggest that the optimal range for $s^{\mathbf{z}}_{\text{until}}$ is 10 to 20, as this balances retaining original details with achieving effective changes. Similarly, the optimal range for $s^{\mathbf{A}}_{\text{until}}$ is approximately 20 to 50. 

\begin{figure}[t]
    \centering
    \includegraphics[width=\linewidth]{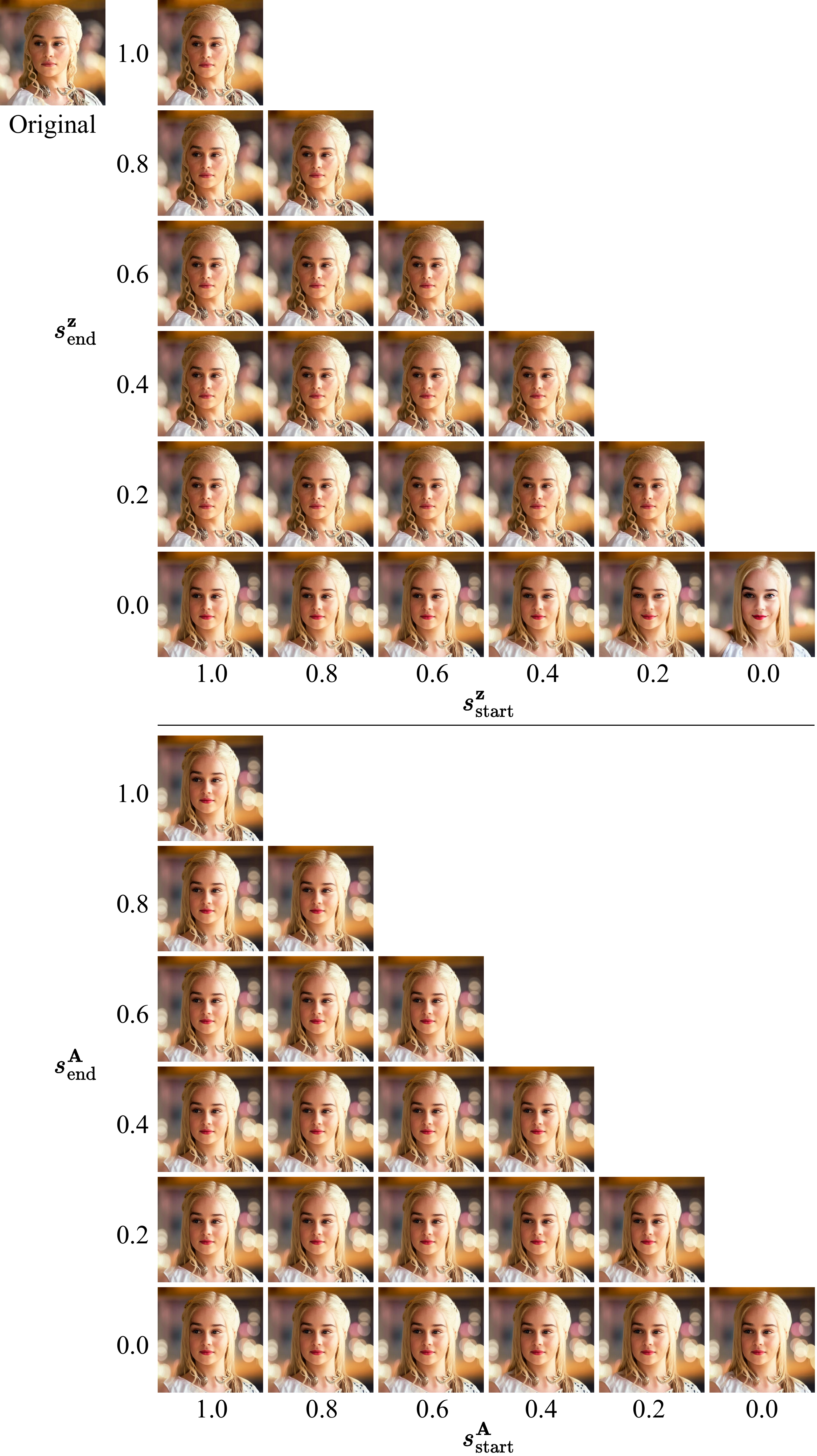}
    \captionof{figure}{Ablation study on scheduler parameters \textbf{Decay start} and \textbf{Decay end}. Results show the effect of varying $(s^{\mathbf{z}}_{\text{start}}, s^{\mathbf{z}}_{\text{end}})$ for latent mixing and $(s^{\mathbf{A}}_{\text{start}}, s^{\mathbf{A}}_{\text{end}})$ for attention mixing.}
    \label{fig:ablation_scheduler_startend_1}
\end{figure}

\begin{figure}[t]
    \centering
    \includegraphics[width=\linewidth]{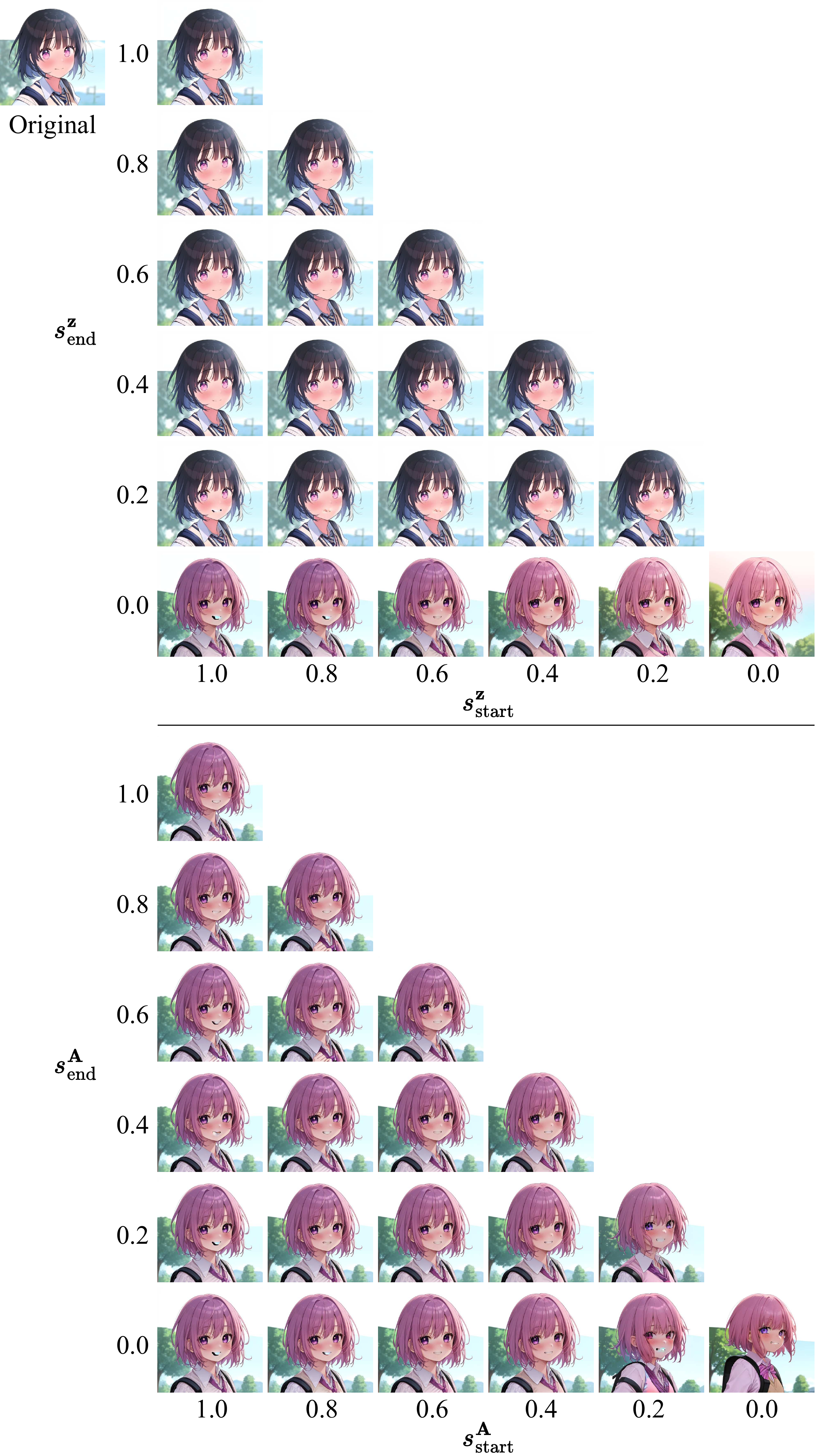}
    \captionof{figure}{Another example from the ablation study on scheduler parameters \textbf{Decay start} and \textbf{Decay end}, demonstrating the impact of different settings for latent and attention mixing.}
    \label{fig:ablation_scheduler_startend_2}
\end{figure}

We also investigated the starting and ending mixing scales in the schedulers, specifically $(s^{\mathbf{z}}_{\text{start}}, s^{\mathbf{z}}_{\text{end}})$ and $(s^{\mathbf{A}}_{\text{start}}, s^{\mathbf{A}}_{\text{end}})$. Since the schedulers follow a decaying pattern, we restrict $s_{\text{start}} \geq s_{\text{end}}$. The results, illustrated in Fig.~\ref{fig:ablation_scheduler_startend_1} and \ref{fig:ablation_scheduler_startend_2}, show that when the ending value for latent mixing is slightly above zero, the results resemble the original image closely, indicating that the integration of latent information is best when it decays near zero. For attention mixing, the differences are minimal as long as $s^{\mathbf{A}}_{\text{start}} \geq 0.4$.

These findings emphasize that latent mixing should be applied more intensively in the early stages of the denoising process to incorporate signals from the original image, with reduced mixing in later steps. Similarly, attention mixing is most effective when applied early to enhance structural preservation. Its impact diminishes in later steps, suggesting that integrating additional attention maps during these stages has minimal effect on the final result.

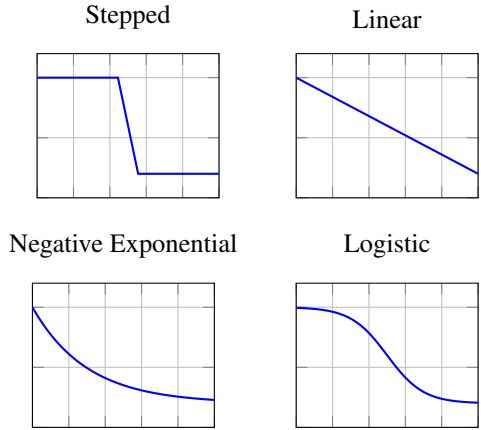
\begin{figure}[t]
    \centering
    \begin{tabular}{cc}
        \begin{tikzpicture}
            \begin{axis}[
                width=4cm, height=3.5cm,
                grid=major,
                xmin=0, xmax=10,
                ymin=0, ymax=1.2,
                title={Stepped},
                xticklabel=\empty,
                yticklabel=\empty,
            ]
            \addplot+[domain=0:10, samples=10, mark=none, thick] {1 - floor(x/5) * (1-0.2)};
            \end{axis}
        \end{tikzpicture}
        &
        \begin{tikzpicture}
            \begin{axis}[
                width=4cm, height=3.5cm,
                grid=major,
                xmin=0, xmax=10,
                ymin=0, ymax=1.2,
                title={Linear},
                xticklabel=\empty,
                yticklabel=\empty,
            ]
            \addplot+[domain=0:10, samples=100, mark=none, thick] {1 - x * (1 - 0.2) / 10};
            \end{axis}
        \end{tikzpicture}
        \\
        \begin{tikzpicture}
            \begin{axis}[
                width=4cm, height=3.5cm,
                grid=major,
                xmin=0, xmax=10,
                ymin=0, ymax=1.2,
                title={Negative Exponential},
                xticklabel=\empty,
                yticklabel=\empty,
            ]
            \addplot+[domain=0:10, samples=100, mark=none, thick] {0.2 + (1 - 0.2) * exp(-x/3)};
            \end{axis}
        \end{tikzpicture}
        &
        \begin{tikzpicture}
            \begin{axis}[
                width=4cm, height=3.5cm,
                grid=major,
                xmin=0, xmax=10,
                ymin=0, ymax=1.2,
                title={Logistic},
                xticklabel=\empty,
                yticklabel=\empty,
            ]
            \addplot+[domain=0:10, samples=100, mark=none, thick] {0.2 + (1 - 0.2) / (1 + exp((x-5)))};
            \end{axis}
        \end{tikzpicture}
    \end{tabular}

    \caption{Comparison of decay functions: stepped, linear, negative exponential, and logistic.}
    \label{fig:decay_functions}
\end{figure}

\begin{figure}[t]
    \centering
    \includegraphics[width=\linewidth]{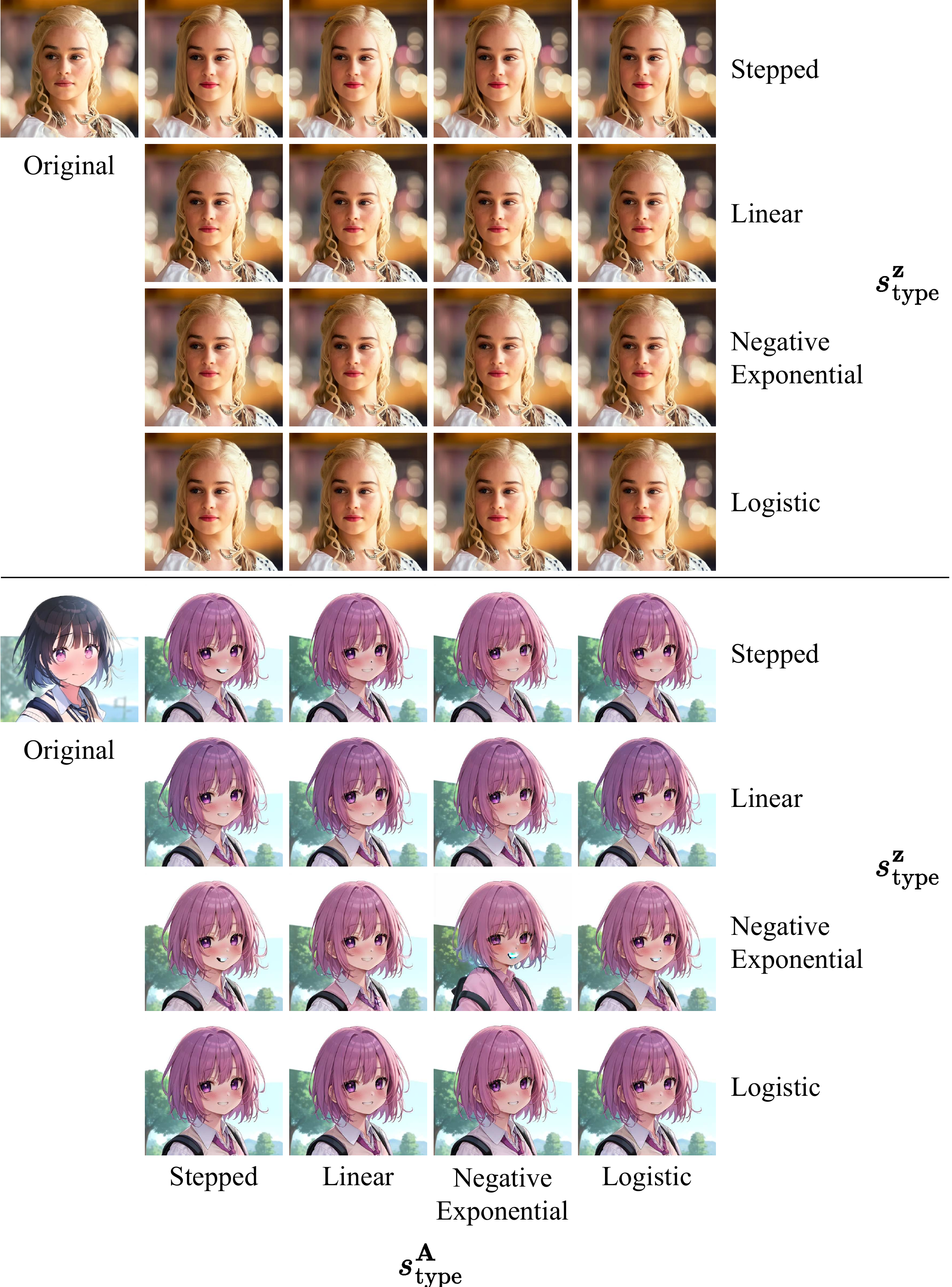}
    \captionof{figure}{Image editing results with different types of schedulers. $s^{\mathbf{A}}_{\text{type}}$ and $s^{\mathbf{z}}_{\text{type}}$ indicate the scheduler types assigned for attention mixing and latent mixing, respectively.}
    \label{fig:ablation_scheduler_type}
\end{figure}

Finally, we compare the results using different scheduler types. We explored four decay functions for LAMS (Fig.~\ref{fig:decay_functions}) to dynamically control mixing proportions: stepped, linear, negative exponential, and logistic. Each function dictates how the mixing scales for latent and attention maps evolve across denoising steps. Stepped decay introduces abrupt changes at predefined points, while linear decay ensures a gradual transition. Negative exponential decay starts with a sharp drop that slows over time, whereas logistic decay follows a smooth S-shaped curve for more gradual adjustments. As shown in Fig.~\ref{fig:ablation_scheduler_type}, the differences between these scheduler types are subtle, with minimal impact on overall image quality. Only minor details, such as hair and clothing, show slight variations. Therefore, the choice of scheduler type is not a critical factor for performance.

\begin{figure}[t]
    \centering
    \includegraphics[width=\linewidth]{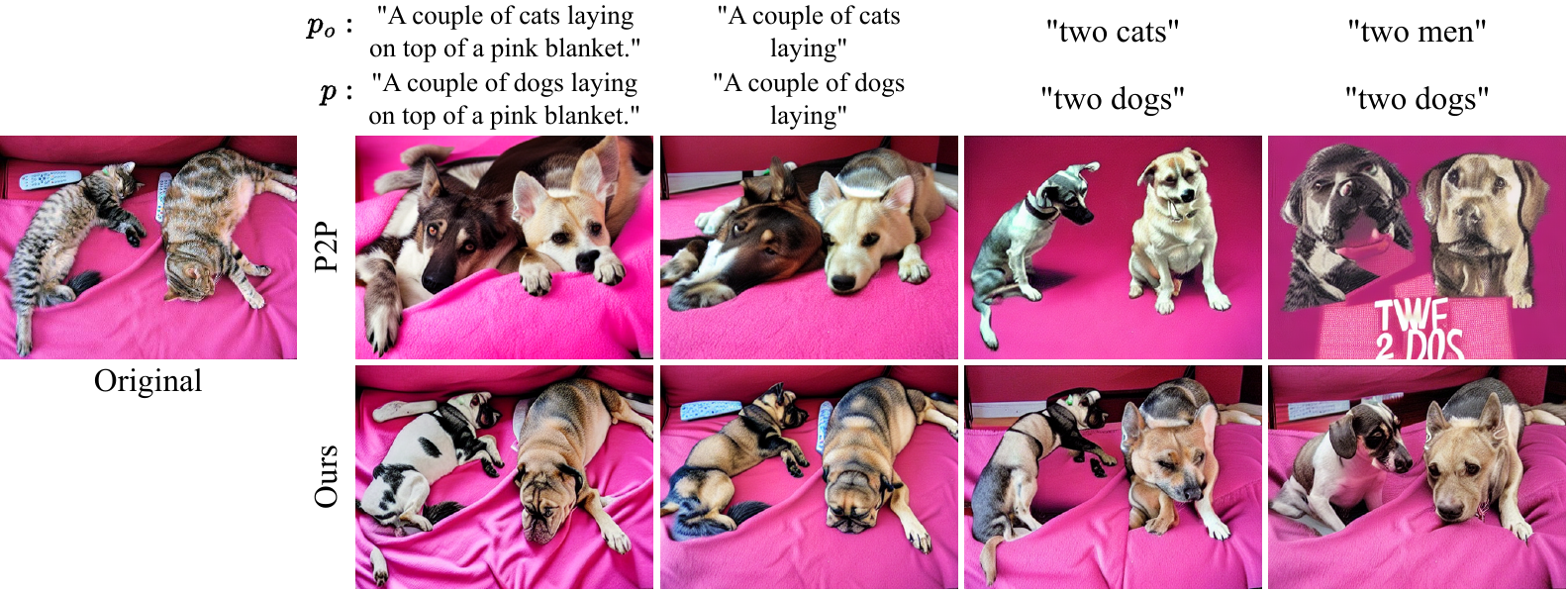}
    \captionof{figure}{Effectiveness of LAMS on differing degrees of alignment between the prompt and the original image.}
    \label{fig:exp_prompt_alignment}
\end{figure}

We further evaluate the effectiveness of LAMS in scenarios where the original prompt $p_o$ poorly aligns with the target image. As shown in Fig.~\ref{fig:exp_prompt_alignment}, we compare our method with the P2P baseline under varying degrees of prompt-image alignment. The results demonstrate that, despite subtle differences, LAMS consistently outperforms P2P across different levels of alignment, particularly in preserving the object integrity specified by the target prompt.

\section{Other Materials}
\label{sup:other_materials}

\subsection{User Study Questionnaire}
\label{sup:user_study_questionnaire}

\begin{figure}[t]
    \centering
    \includegraphics[width=0.9\linewidth]{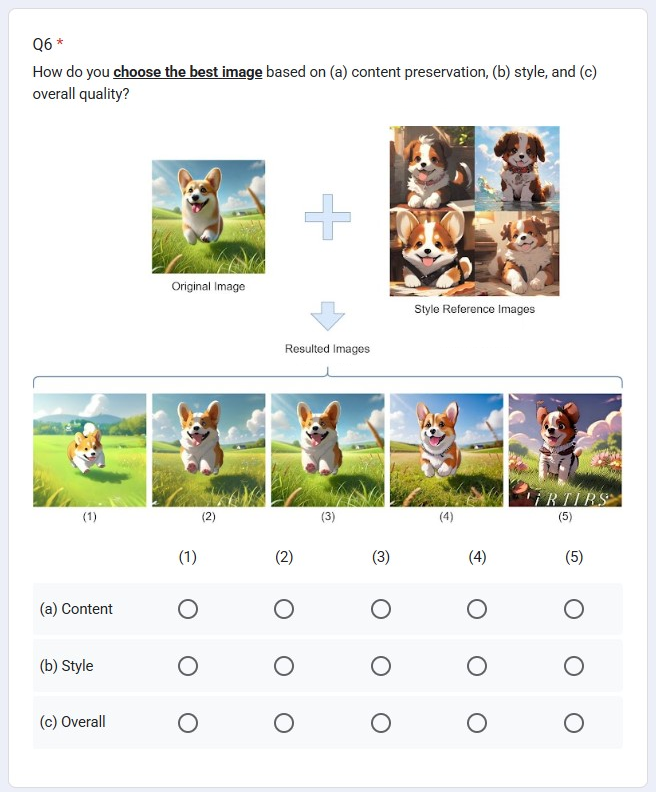}
    \captionof{figure}{A screenshot of the questionnaire for style transfer user study.}
    \label{fig:questionnaire_screenshot}
\end{figure}

Since style transfer is hard to evaluate quantitatively, we conducted a user study comparing five approaches—including our method with and without masking. For 15 images each subjected to a different style transfer, 41 participants were shown both the original and the transformed images and asked to vote on which method was superior in terms of content preservation, style application, and overall quality. Figure~\ref{fig:questionnaire_screenshot} shows a screenshot of one of the 15 questions in the questionnaire created using Google Forms for the style transfer evaluation. The instructions provided to participants are shown in Box~\ref{box:user_study_instructions}.

\begin{instructionbox}[label=box:user_study_instructions]{Participant Instructions for User Study}
\textbf{Please read these instructions carefully. In this study, you will see:}
\begin{itemize}
    \item \textbf{Original Image:} The original content.
    \item \textbf{Style Reference Images:} The artistic style to apply.
    \item \textbf{Resulted Images:} 5 versions of the original image with different styles applied.
\end{itemize}

\textbf{Your task is to evaluate these 5 images based on:}
\begin{itemize}
    \item \textbf{(a) Content Preservation:} How well does the image keep the original content (e.g., character identity, background, shape, etc.)?
    \item \textbf{(b) Style:} How well does the image apply the artistic style? (Does the style look like the reference style images?)
    \item \textbf{(c) Overall:} Which image do you prefer overall?
\end{itemize}

\textbf{Select one image for each category.}
\end{instructionbox}

\end{document}